\documentclass[runningheads]{llncs}

\usepackage{eccv}

\usepackage{eccvabbrv}

\usepackage{graphicx}
\usepackage{booktabs}

\usepackage[accsupp]{axessibility}  

\usepackage[pagebackref,breaklinks,colorlinks,citecolor=eccvblue]{hyperref}

\usepackage{orcidlink}

\usepackage{amsmath}
\usepackage{amsfonts}
\usepackage{amssymb}
\usepackage{graphicx}
\usepackage{multirow}
\usepackage{caption}
\usepackage[table]{xcolor}
\usepackage{arydshln}
\usepackage{pifont}
\usepackage{subcaption}
\usepackage{xcolor}

\usepackage{multibib}
\newcites{supp}{Supplementary References}

\begin{document}

\title{{\scshape SFKD}: Spatial--Frequency Joint-Aware Heterogeneous Knowledge Distillation via Multi-Level Wavelet Spectral Interaction} 

\titlerunning{SFKD}

\author{Cuipeng Wang\inst{1,2}\orcidlink{0009-0009-6299-4729} \and
Haipeng Wang\inst{1,2}\orcidlink{0000-0003-1912-7143}\thanks{Corresponding author.}}

\authorrunning{C.~Wang et al.}


\institute{
Key Laboratory for Information Science of Electromagnetic Waves,\\
Ministry of Education, Fudan University, Shanghai, China\\
\and
Discipline and Technology Center of Microwave Vision Intelligent Sensing,\\
Fudan University, Shanghai, China\\
\email{cpwang23@m.fudan.edu.cn, hpwang@fudan.edu.cn}
}

\maketitle

\begin{abstract}
Most existing knowledge distillation methods focus on homogeneous models (\eg, CNN$\rightarrow$CNN), thereby overlooking the flexibility and potential of knowledge transfer across heterogeneous models.
Due to intrinsic inductive bias discrepancies between heterogeneous models that cause spatial distribution inconsistencies, prior heterogeneous distillation methods often weaken or discard spatial information in heterogeneous representations.
However, the spatial information in representations often encodes transferable global structural semantics as well as architecture-specific local details, and therefore should not be directly ignored.
To better leverage the spatial information encoded in heterogeneous representations, we propose a Spatial--Frequency Joint-Aware Heterogeneous Knowledge Distillation framework (SFKD).
By leveraging the complementary properties of wavelet transform spatial locality and Fourier representations in characterizing global energy distributions, we first apply multi-level discrete wavelet transform to explicitly decouple spatial information. 
The resulting wavelet sub-bands are further refined by a dual-stream dual-stage refinement module, and finally combined with a Gaussian-filtered frequency loss to selectively capture informative global information.
Extensive experiments on multiple benchmark datasets under both homogeneous and heterogeneous models demonstrate the superiority of our method.
Code is available at \url{https://github.com/cpcpWang/SFKD}.

\end{abstract}

\begin{figure}[t]
\centering
\includegraphics[width=0.8\textwidth]{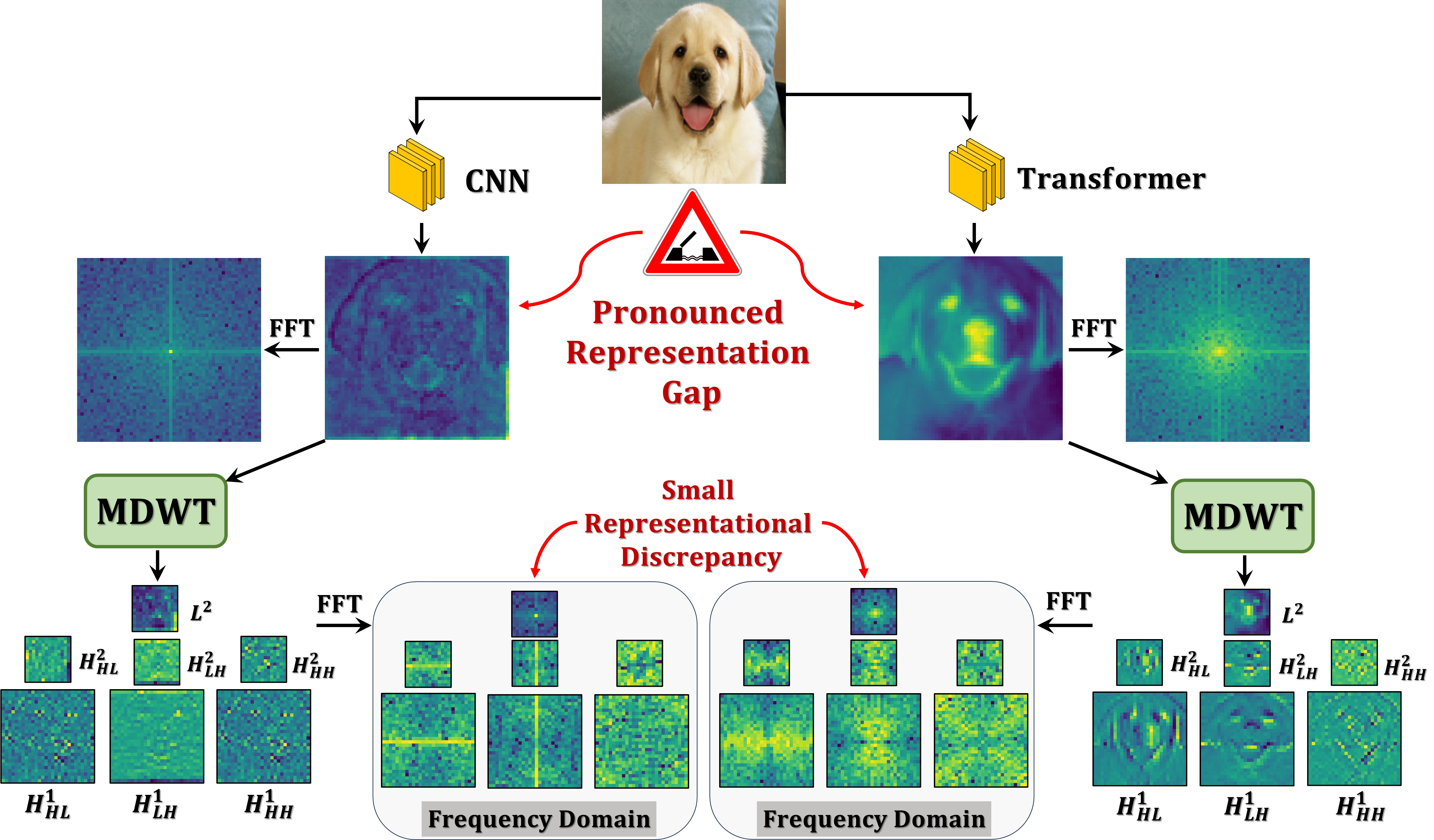}
\caption{\textbf{Illustration of heterogeneous feature maps, multi-level wavelet sub-bands, and corresponding FFT spectral features.} Heterogeneous feature maps exhibit substantial spatial discrepancies, while the wavelet sub-bands obtained through MDWT are further transformed by FFT, resulting in spectral features with reduced discrepancies.}
\label{fig1}
\vskip -0.35in
\end{figure}
\begin{figure}[t]
\centering
\includegraphics[width=\textwidth, height=6.5cm]{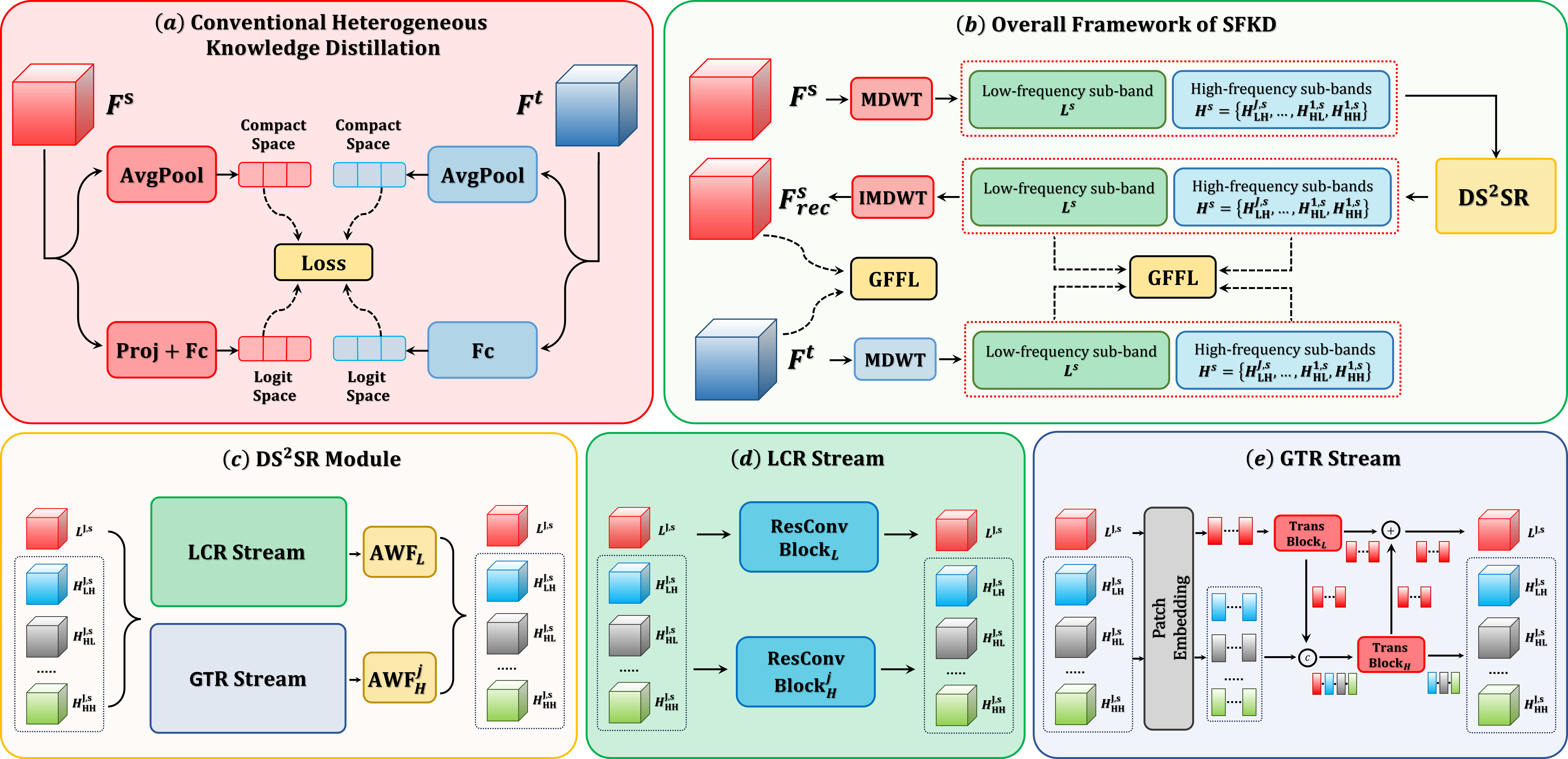}
\caption{\textbf{Comparison between conventional heterogeneous distillation (a) and the proposed SFKD (b).} (c) Detailed architecture of the DS$^2$SR module. The LCR stream and GTR stream are illustrated in (d) and (e), respectively. Unlike conventional methods that directly compress or suppress spatial information, SFKD explicitly decouples spatial information via MDWT, further refines wavelet sub-bands with DS$^2$SR, and finally performs frequency-domain alignment using GFFL, thereby enabling more effective exploitation of spatial information in heterogeneous representations.}
\label{fig2}
\vskip -0.35in
\end{figure}
\vspace{-20pt}
\section{Introduction}
\label{sec1}
With the rapid development of sophisticated architectures such as CNNs~\cite{he2016deep,sandler2018mobilenetv2,liu2022convnet}, Transformers~\cite{tolstikhin2021mlp,touvron2022resmlp}, and MLP-based models~\cite{liu2021swin,dosovitskiy2020image,touvron2021training}, deep learning has achieved remarkable success in computer vision.
However, the continuous increase in model parameters, while improving performance, also incurs substantial computational and memory overhead, posing significant challenges for deployment on resource-constrained edge devices.
Knowledge Distillation (KD) has been widely recognized as an effective technique for model compression. 
The core idea is to transfer knowledge from a powerful teacher model to a compact student model, thereby improving the student’s performance without increasing its model size.

Most existing knowledge distillation methods~\cite{hinton2015distilling,huang2022knowledge,zhao2022decoupled,romero2014fitnets,tian2019contrastive,chen2021distilling,heo2019comprehensive,park2019relational,peng2019correlation,liu2023function} primarily focus on knowledge transfer between homogeneous models. 
However, merely restricting distillation to homogeneous models introduces several limitations. 
On the one hand, for a given student model, high-performing homogeneous teacher models suitable for distillation are often scarce, making the selection of an optimal teacher non-trivial. 
On the other hand, restricting distillation to homogeneous models may fail to exploit the complementary inductive biases inherent in heterogeneous models.
Recent studies on hybrid architectures~\cite{li2023uniformer,li2023convmlp} have shown that combining heterogeneous modules, such as CNNs and Transformers, can better leverage complementary strengths and yield superior performance.
Consequently, relying solely on homogeneous teachers may prevent the student model from benefiting from such cross-architecture complementary knowledge, resulting in suboptimal distillation performance.
For instance, as shown in \cref{table2} and \cref{table3}, with our proposed method, distilling a ResNet18 student from a heterogeneous Swin-T teacher achieves better performance (72.54\%) than using a homogeneous ResNet34 teacher (72.34\%).
This observation further highlights the potential advantages of cross-architecture distillation.
Motivated by these insights, this work focuses on knowledge distillation across heterogeneous models, aiming to enlarge the teacher selection space and enhance distillation flexibility by leveraging complementary cross-architecture knowledge.

Hao et al.~\cite{hao2023one} use CKA to analyze intermediate representations across heterogeneous architectures and reveal significant discrepancies between them.
Such discrepancies primarily arise from intrinsic inductive bias differences across heterogeneous models~\cite{park2022vision, raghu2021vision}.
CNNs~\cite{he2016deep,sandler2018mobilenetv2,liu2022convnet} exhibit strong locality and translation invariance due to convolutional operations, whereas Transformers~\cite{liu2021swin,dosovitskiy2020image,touvron2021training} and MLP-based models~\cite{tolstikhin2021mlp,touvron2022resmlp} rely on patchification and long-range dependency modeling, leading to distinct structural properties.

Bridging this representational gap is crucial for effective knowledge distillation between heterogeneous models.
Prior studies make significant efforts in this direction.
However, we observe that, due to the pronounced spatial distribution discrepancies between heterogeneous representations, as illustrated in \cref{fig1}, several approaches often mitigate this gap by reducing explicit spatial dependencies, thereby limiting the effective utilization of spatial structures.
For example, Hao et al.~\cite{hao2023one} maps intermediate features to the logit space to alleviate representational differences, Wu et al.~\cite{wu2024aligning} align representations in a compact embedding space, and Li et al.~\cite{li2025fuse} introduces a hybrid bridging model with a structure-agnostic loss.
However, spatial information in intermediate representations inherently encode both transferable global semantics and architecture-specific local details.
Excessively discarding or suppressing such spatial information may limit the effective utilization of complementary cross-architecture knowledge.

To better exploit the spatial information encoded in representations of heterogeneous architectures, we propose a Spatial--Frequency Joint-Aware Heterogeneous Knowledge Distillation framework (\textbf{SFKD}).
Specifically, we first apply multi-level discrete wavelet transform to explicitly decouple teacher and student representations in the spatial domain.
In the resulting wavelet spectrum, the low-frequency(LF) sub-band primarily captures global structural semantics, while the high-frequency(HF) sub-bands encode local details such as edges and textures.
Due to the relatively limited representational capacity of the student network, its wavelet sub-bands often exhibit greater structural instability and distribution discrepancies compared to those of the teacher, which hinders effective feature alignment.
To address this issue, we introduce a novel dual-stream dual-stage refinement module to further refine the student wavelet sub-bands, enhancing their structural consistency.
Considering the spatial discrepancies between heterogeneous representations, directly aligning sub-bands in the spatial domain may be unstable.
In contrast, the Fourier domain can characterize global energy distribution and overall structural patterns.
Therefore, we design a Gaussian-filtered frequency loss to selectively emphasize the dominant information in each wavelet sub-band, facilitating effective heterogeneous knowledge transfer.
Wavelet decomposition provides spatial locality, whereas Fourier transformation offers global spectral modeling.
By integrating these complementary properties, SFKD establishes a spatial–frequency joint-aware distillation framework that effectively extracts and emphasizes informative spatial information in heterogeneous representations. 
Extensive experiments on multiple benchmark datasets demonstrate that SFKD consistently outperforms existing methods and achieves state-of-the-art or competitive performance across diverse settings.

In summary, the main contributions of our work are as follows:
\begin{itemize}
\item We provide a systematic analysis of existing heterogeneous distillation methods and identify their limited exploitation of spatial information in intermediate representations.
To address this issue, we propose to integrate multi-level wavelet transform with Fourier-domain modeling to decouple, filter, and effectively transfer informative spatial information.
\item We propose a Spatial--Frequency Joint-Aware Heterogeneous Knowledge Distillation framework (SFKD).
The framework applies multi-level discrete wavelet transform to explicitly decouple spatial information in heterogeneous representations, incorporates a dual-stream dual-stage refinement module to enhance student wavelet sub-bands, and then employs a Gaussian-filtered frequency loss to facilitate efficient heterogeneous knowledge transfer.
\item Extensive experiments on CIFAR-100 and ImageNet-1K under both heterogeneous and homogeneous settings demonstrate the superiority of SFKD in improving student performance, while also validating the effectiveness of decoupling and selectively exploiting spatial information in representations.
\end{itemize}
\begin{figure}[t]
\centering
\begin{subfigure}[b]{0.3\textwidth}
\centering
\includegraphics[width=\linewidth,height=6.5cm]{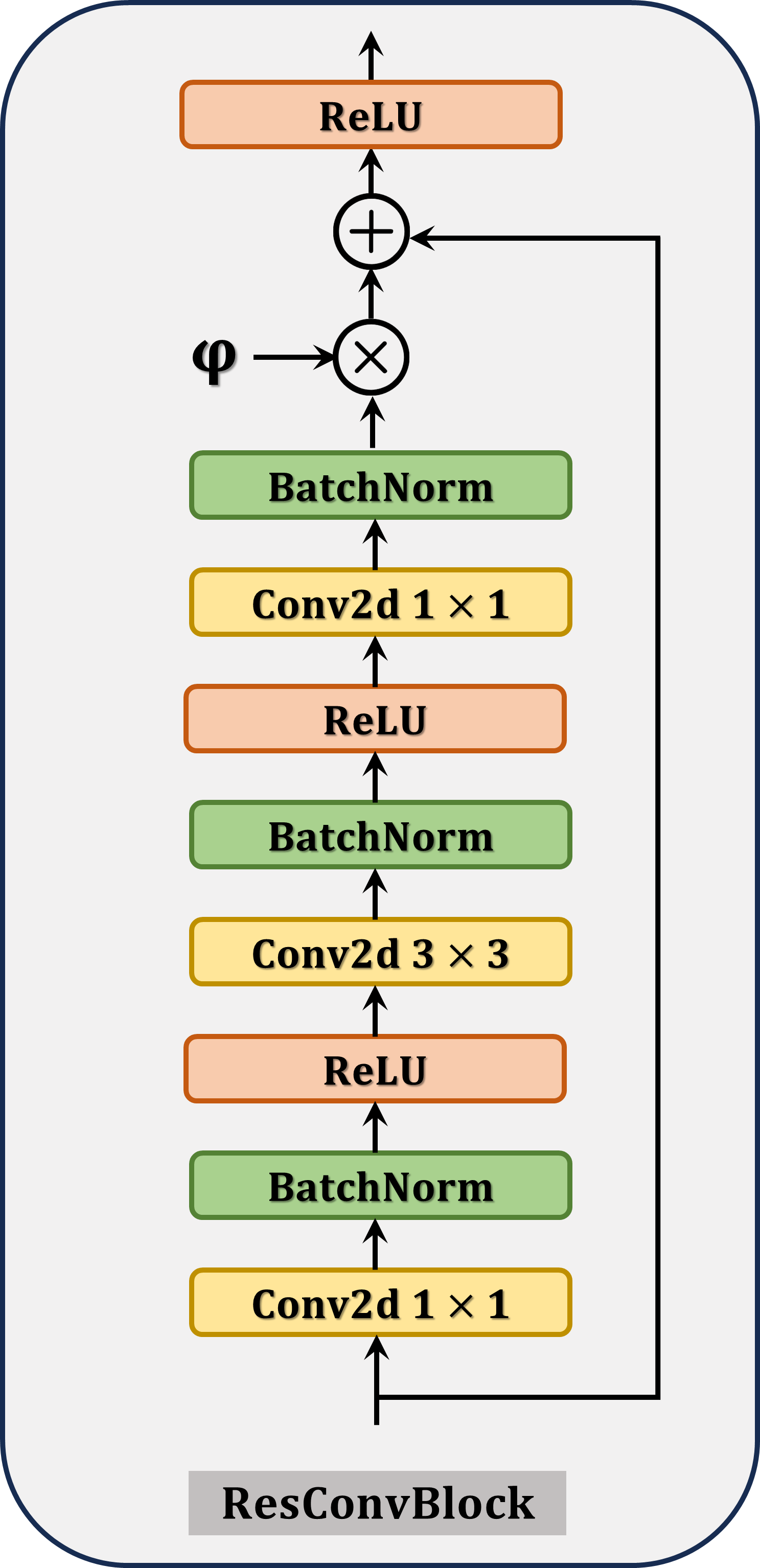}
\caption{ResConvBlock}
\label{fig3-1}
\end{subfigure}
\hfill
\begin{subfigure}[b]{0.3\textwidth}
\centering
\includegraphics[width=\linewidth,height=6.5cm]{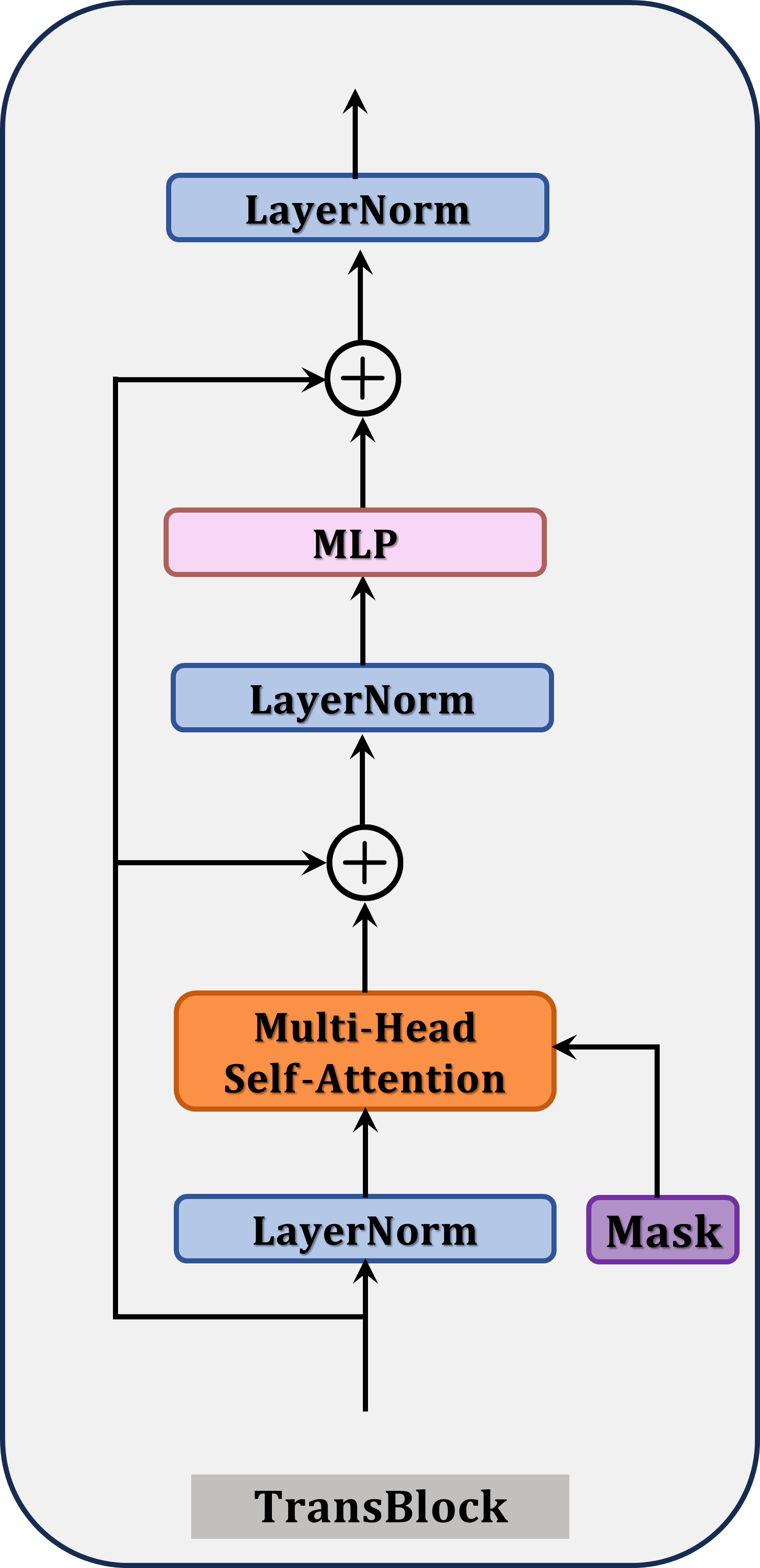}
\caption{TransBlock}
\label{fig3-2}
\end{subfigure}
\hfill
\begin{subfigure}[b]{0.3\textwidth}
\centering
\includegraphics[width=\linewidth,height=6cm]{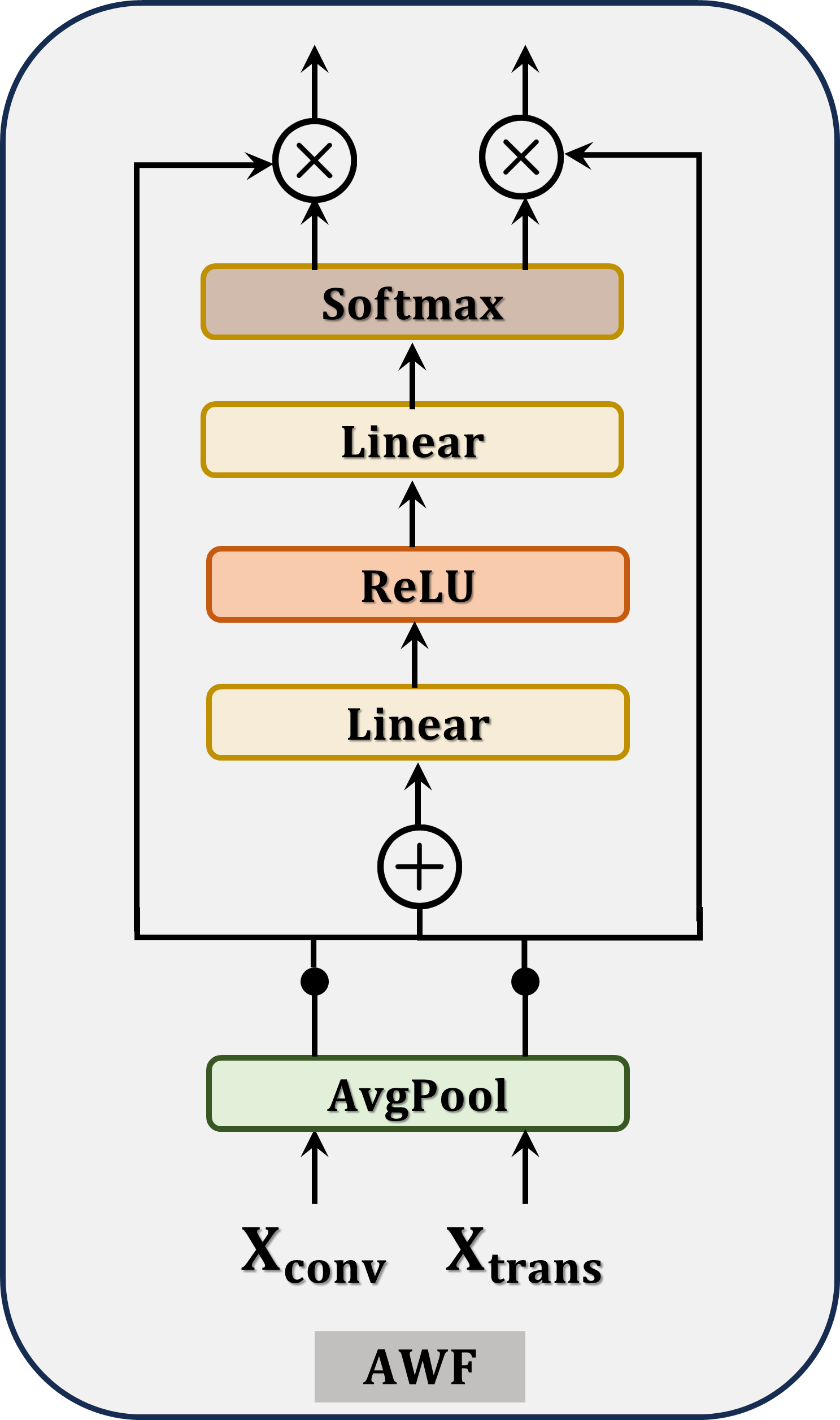}
\caption{AWF}
\label{fig3-3}
\end{subfigure}
\caption{\textbf{The architectural details of the ResConvBlock, TransBlock, and Adaptive Weighted Fusion (AWF) modules.}}
\label{fig3}
\vskip -0.25in
\end{figure}
\section{Related Work}
\label{sec2}
We discuss \textbf{Related Work} and defer a concentrated account to the Appendix (see \cref{appendix1}). 
\section{Method}
\label{sec3}
\subsection{Overall Framework}
\label{sec3-1}
As illustrated in \cref{fig2}, the proposed SFKD framework primarily consists of three core components: the Multi-level Discrete Wavelet Transform (\textbf{MDWT}) module, the Dual-Stream Dual-Stage Spectral Refinement (\textbf{DS$^2$SR}) module, and the Gaussian-Filtered Frequency Loss (\textbf{GFFL}).
It should be noted that the student features $\mathbf{F}^s$ mentioned below refer to the representations that have been projected by a projector and aligned with the teacher features $\mathbf{F}^t$. 
For simplicity, this projection process is omitted in \cref{fig2} and the following descriptions.

Compared with existing heterogeneous distillation methods~\cite{hao2023one,li2025fuse,wu2024aligning} that typically weaken or discard spatial information in intermediate representations, we instead explicitly model such spatial characteristics in a structured manner.
Specifically, we first apply a MDWT to both teacher and student representations, facilitating an explicit spatial decoupling of intermediate representations into LF and HF sub-band components. 
Considering the relatively limited representation capacity of the student network, its wavelet-decomposed sub-bands tend to exhibit greater structural instability and representational discrepancies compared to those of the teacher.  
Such discrepancies hinder effective alignment between student and teacher representations. 
To mitigate these discrepancies, we propose a novel dual-stream dual-stage spectral refinement (DS$^2$SR) module that adaptively refines the student's wavelet sub-bands, recovering structural details and global semantic consistency degraded in the student network.
In \cref{sec4-3-1}, we validate the necessity of sub-band refinement and the effectiveness of the proposed DS$^2$SR module through ablation studies.
Moreover, motivated by the efficacy of Fourier spectra representations in characterizing stable characterization of global energy distribution and structural information, we propose a Gaussian-Filtered Frequency Loss (GFFL).
By incorporating Gaussian-filtered mask, GFFL selectively emphasizes informative frequency components within each sub-band, thereby facilitating effective knowledge transfer and enabling joint spatial–frequency awareness in the distillation process.
\begin{figure}[t]
\centering
\begin{subfigure}[b]{0.3\textwidth}
\centering
\includegraphics[width=\linewidth]{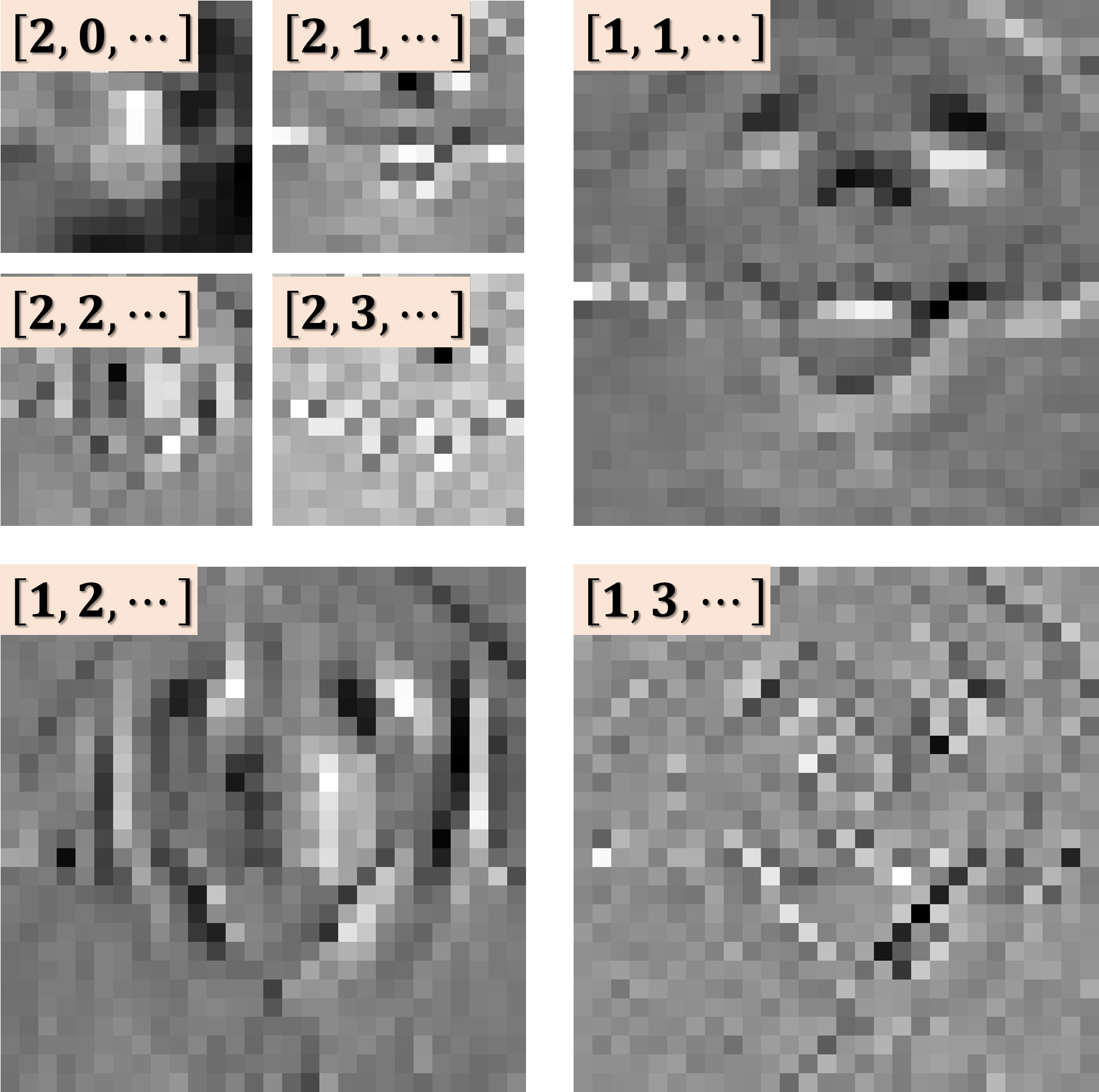}
\caption{4D positional encoding}
\label{fig4-1}
\end{subfigure}
\hfill
\begin{subfigure}[b]{0.2\textwidth}
\centering
\includegraphics[width=\linewidth]{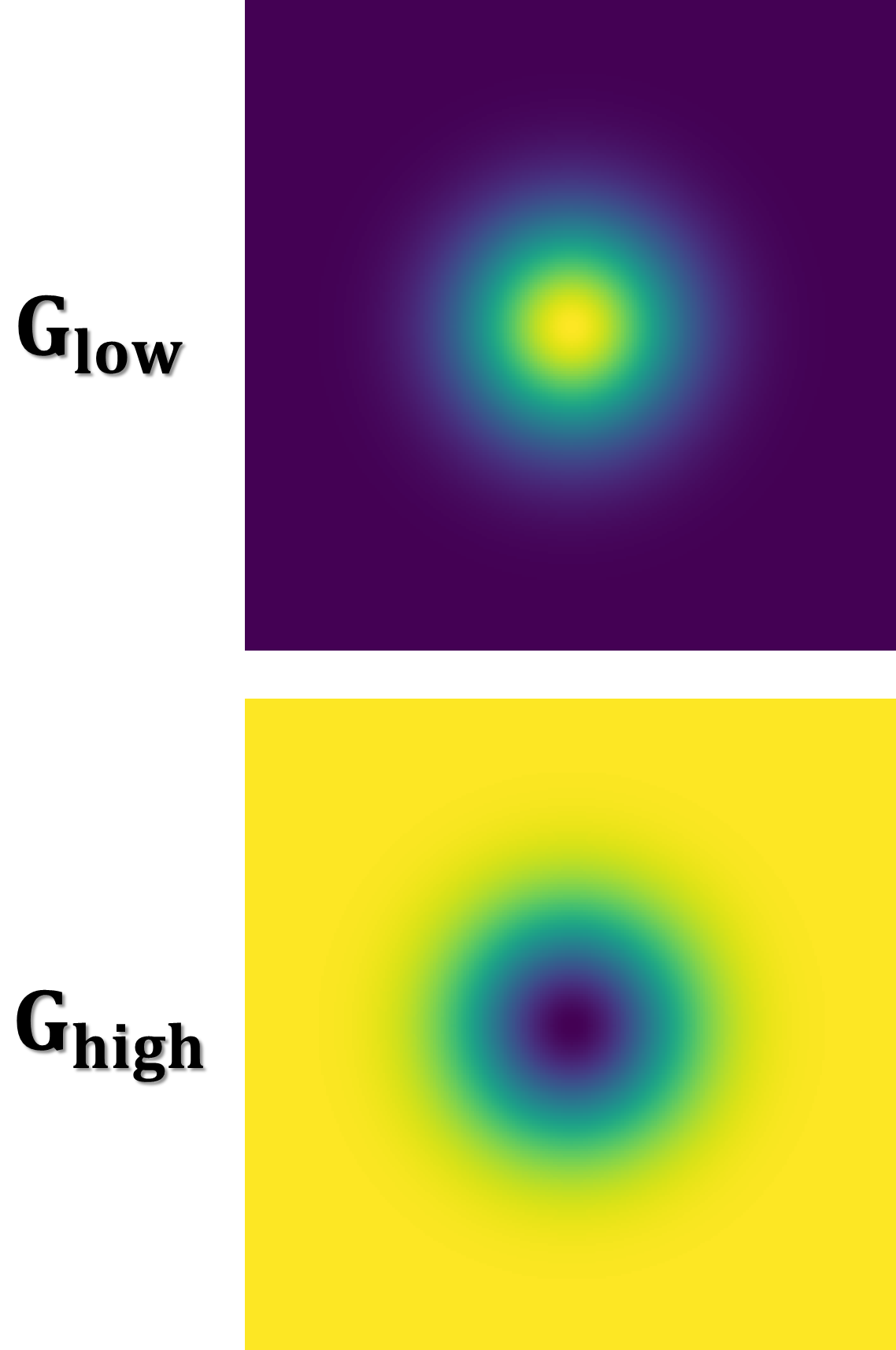}
\caption{Gaussian filters}
\label{fig4-2}
\end{subfigure}
\hfill
\begin{subfigure}[b]{0.3\textwidth}
\centering
\includegraphics[width=\linewidth]{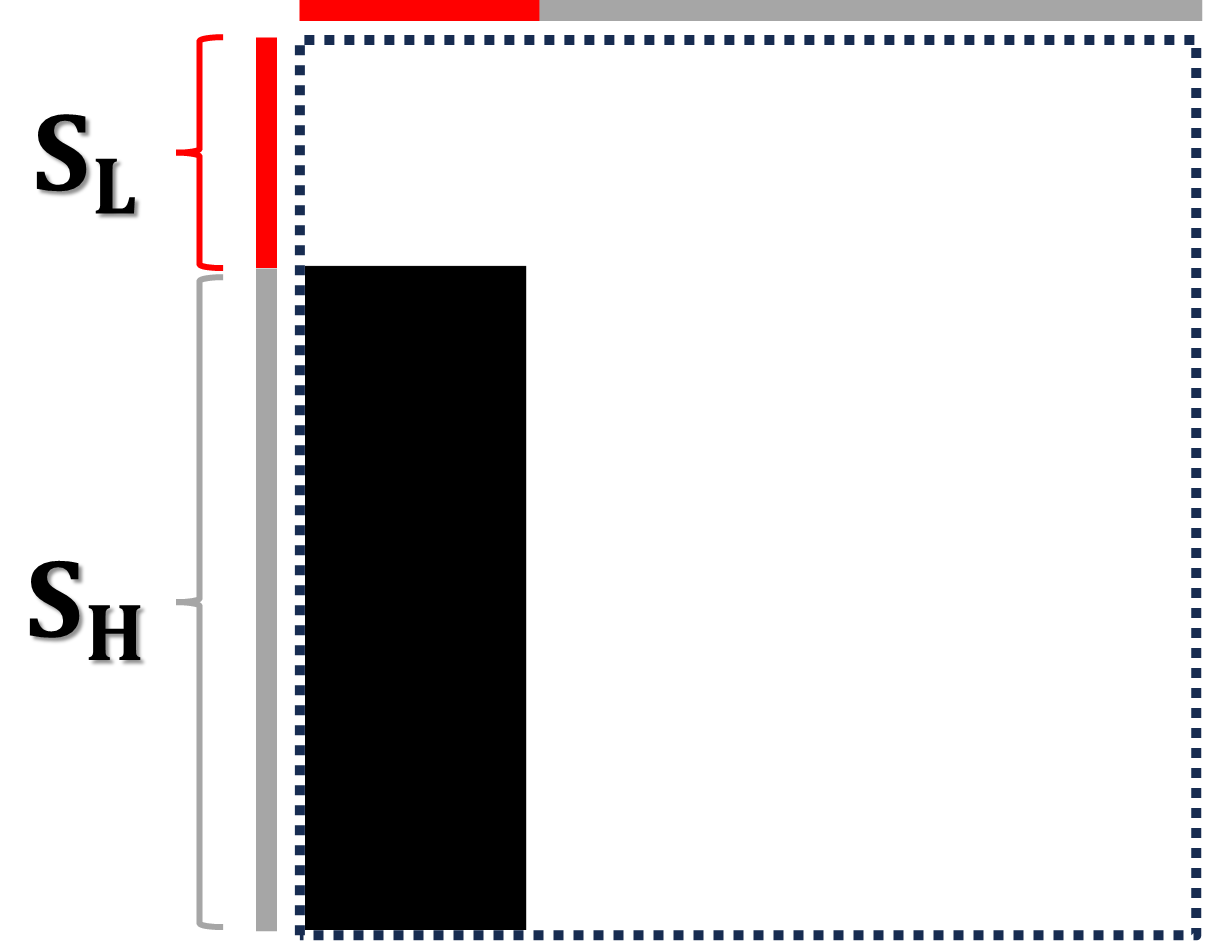}
\caption{Attention mask}
\label{fig4-3}
\end{subfigure}
\caption{\textbf{Illustration of the proposed 4D positional encoding, Gaussian low-pass and high-pass filtering masks, and the designed attention mask.} In \cref{fig4-3}, the \textcolor{red}{red lines} denote low-frequency tokens, while the \textcolor{gray}{gray lines} denote high-frequency tokens. }
\label{fig4}
\vskip -0.3in
\end{figure}
\subsection{Multi-level Discrete Wavelet Transform (\textbf{MDWT})}
\label{Sec3-2}
Wavelet transform is a general signal decomposition technique that explicitly decouples low-frequency and high-frequency components within representations.
In this work, we adopt the Haar wavelet transform~\cite{haar1909theorie} as the underlying decomposition operator.
Given an input feature tensor $\mathbf{F} \in \mathbb{R}^{B \times C \times H \times W}$, we apply the discrete wavelet transform operator $\mathrm{DWT}(\cdot)$ independently to each sample in the batch to obtain one low-frequency sub-band $\mathbf{L}^{1} \in \mathbb{R}^{B \times C \times \frac{H}{2} \times \frac{W}{2}}$ and three high-frequency sub-bands $\mathcal{H}^{1}=\left\{
\mathbf{H}_{LH}^{1},\mathbf{H}_{HL}^{1},\mathbf{H}_{HH}^{1}\right\}\in\mathbb{R}^{B \times C\times \frac{H}{2} \times \frac{W}{2}},$
corresponding to vertical, horizontal, and diagonal frequency components, respectively.
\begin{equation}
(\mathbf{L}^{1}, \mathcal{H}^{1}) = \mathrm{DWT}(\mathbf{F})
\label{eq1}
\end{equation}

By recursively applying the discrete wavelet transform to the low-frequency sub-band for $J$ levels, we obtain the $J$-th level wavelet decomposition, which consists of a low-frequency sub-band $\mathbf{L}^{J}\in
\mathbb{R}^{B \times C \times \frac{H}{2^{J}} \times \frac{W}{2^{J}}}$ and a high-frequency sub-band set $\mathcal{H}^{J}\in
\mathbb{R}^{B \times C \times \frac{H}{2^{J}} \times \frac{W}{2^{J}}}$.

After performing a $J$-level discrete wavelet transform on the input feature map 
$\mathbf{F}$, we obtain the complete set of multi-level sub-bands, denoted as $\mathbf{X}$,
\begin{equation}
\mathbf{X} = \left\{
\mathbf{L}^{J},
\mathcal{H}^{J},
\mathcal{H}^{J-1},
\ldots,
\mathcal{H}^{1}
\right\}.
\label{eq2}
\end{equation}

We collectively denote these sub-bands as the $J$-level wavelet spectrum of $\mathbf{F}$, defined as:
\begin{equation}
\mathbf{X} = \mathrm{MDWT}(\mathbf{F}, J).
\label{eq3}
\end{equation}

Since the discrete wavelet transform is inherently invertible, the original feature map $\mathbf{F}$ can be reconstructed through the corresponding $J$-level inverse $\mathrm{DWT}$, defined as:
\begin{equation}
\mathbf{F} = \mathrm{IMDWT}(\mathbf{X}, J).
\label{eq4}
\end{equation}
\subsection{Dual-Stream Dual-Stage Spectral Refinement(DS$^2$SR)}
\label{sec3-3}
Although the discrete wavelet transform explicitly decouples low- and high-frequency components, it does not inherently eliminate representational noise and structural instability.
Moreover, since practical wavelet filters are implemented with finite-length kernels, they cannot achieve ideal band-limited separation~\cite{mallat1999wavelet, strang1996wavelets}. 
As a result, spectral leakage or residual frequency overlap across sub-bands may arise during multi-level decomposition~\cite{unser1997ten, vetterli1995wavelets}.
Consequently, several recent works~\cite{du2025diffusion,li2024ewt,yao2022wave,wang2025wavefusion,finder2024wavelet,huang2024wavedm, miao2024waveface} introduce additional refinement mechanisms after wavelet decomposition to enhance sub-band stability and discriminative capability.
Inspired by these pioneering works and considering the inherent representation discrepancies arising from the divergent inductive biases of heterogeneous models, we propose a novel dual-stream dual-stage spectral refinement (DS$^2$SR) module tailored for heterogeneous knowledge distillation.

Convolutional and self-attention modules possess distinct and highly complementary modeling capabilities~\cite{park2022vision,raghu2021vision}.
Convolutional layers encode strong locality inductive bias, enabling the extraction of fine-grained patterns such as edges and textures, whereas self-attention mechanisms model long-range dependencies through multi-head attention, promoting globally coherent structural representations.
To jointly leverage the complementary modeling capabilities of convolution and self-attention, while enabling architecture-agnostic refinement across heterogeneous representations, we design an novel and adaptive DS$^2$SR module to effectively refine wavelet sub-bands across heterogeneous architectures.
The proposed DS$^2$SR module consists of two functionally distinct yet complementary branches: a Local Convolutional Refinement (\textbf{LCR}) stream and a Global Transformer Refinement (\textbf{GTR}) stream.
The LCR stream consists of convolutional layers and focuses on refining fine-grained local structures within each wavelet sub-band. 
In contrast, the GTR stream leverages self-attention to capture long-range dependencies, thereby enhancing global structural coherence and contextual modeling.
Each branch independently refines the wavelet sub-bands.
As illustrated in \cref{fig1}, considering the distinct statistical properties and structural patterns of low-frequency and high-frequency components, we separately feed the refined low-frequency sub-band and the refined high-frequency sub-band set from both branches into distinct Adaptive Weighted Fusion (\textbf{AWF}) modules for frequency-specific integration.
The detailed architectures of the LCR stream, the GTR stream, and the AWF module are introduced below.
\vspace{-10pt}
\subsubsection{Local Convolutional Refinement (LCR) Stream.}
\label{sec3-3-1}
Due to the distinct statistical distributions and structural characteristics of low- and high-frequency wavelet sub-bands, we apply two independent refinement stages to the low-frequency sub-band and the high-frequency sub-band set to enhance fine-grained local structural modeling.
The detailed architecture of the ResConvBlock is illustrated in \cref{fig3-1}.
ResConvBlock integrates convolutional operations with different strides and a weighted residual connection mechanism.
By fusing the convolutionally refined features with the original input through weighted residual aggregation, it enhances local structural representations while preserving shallow feature information, thereby avoiding over-smoothing and structural degradation.

Specifically, after applying MDWT, the wavelet spectrum is organized into the low-frequency sub-band $\mathbf{L}^{J}$ and and the high-frequency sub-band set $\left\{
\mathcal{H}^{j}\right\}_{j=1}^{J}$.

For the low-frequency sub-band $\mathbf{L}^{J}$, We feed it into a dedicated residual convolutional block ($\mathrm{ResConvBlock}_{L}$) for refinement, yielding the refined low-frequency sub-band denoted as:
\begin{equation}
\mathbf{L}^{J}_{\mathrm{conv}} 
= 
\mathrm{ResConvBlock}_{L}\!\left(\mathbf{L}^{J}\right).
\label{eq5}
\end{equation}

For the high-frequency sub-band set $\{\mathcal{H}^{j}\}_{j=1}^{J}$, 
each level is independently processed by a corresponding residual convolutional block ($\mathrm{ResConvBlock}_{H}^{j}$), yielding:
\begin{equation}
\mathcal{H}^{j}_{\mathrm{conv}} 
= 
\mathrm{ResConvBlock}_{H}^{j}\!\left(\mathcal{H}^{j}\right),
\quad j \in \{1,\dots,J\}.
\label{eq6}
\end{equation}
The final convolutional refinement output for the high-frequency components is denoted as $\mathcal{H}_{\mathrm{conv}} = \{\mathcal{H}^{j}_{\mathrm{conv}}\}_{j=1}^{J}$.
\vspace{-10pt}
\subsubsection{Global Transformer Refinement (GTR) Stream.}
\label{sec3-3-2}
While the LCR stream effectively captures fine-grained local structures, its strong locality inductive bias and limited receptive field restrict its capacity to model long-range dependencies and global contextual information. 
To overcome this limitation, we introduce a Transformer-based global refinement stream that leverages self-attention to aggregate global context and capture long-range structural dependencies in the wavelet sub-bands.

Before feeding the wavelet sub-bands into the $\mathrm{TransBlock}$ for refinement, we first perform tokenization for each level of the wavelet spectrum.
Specifically, the spectrum is separated into the low-frequency sub-band $\mathbf{L}^{J}$ and and the multi-level high-frequency sub-band set $\left\{
\mathcal{H}^{j}\right\}_{j=1}^{J}$.
For each level $j$, the corresponding high-frequency sub-band set $\mathcal{H}^{j}$ is embedded through a convolutional projection layer.
For the low-frequency sub-band $\mathbf{L}^{J}$, a dedicated convolutional projection layer is employed.
The resulting tokens are denoted as $\mathbf{S}_{L}^{J}$ and $\mathcal{S}_{H}
=\left\{\mathbf{S}_{H}^{j}\right\}_{j=1}^{J}$, respectively.
\begin{equation}
\mathbf{S}_{L}^{J} =\mathrm{Conv2d}_{L}\!\left(\mathbf{L}^{J}\right)
\label{eq7}
\end{equation}
\vskip -0.1in
\begin{equation}
\mathbf{S}_{H}^{j}
=
\mathrm{Conv2d}_{H}^{j}\!\left(\mathcal{H}^{j}\right),
\quad j = 1,\dots,J
\label{eq8}
\end{equation}

After convolutional embedding of the wavelet sub-bands, we further introduce positional encoding to preserve structural and hierarchical information.
In standard Vision Transformers~\cite{dosovitskiy2020image}, each image patch is assigned a 2D absolute positional encoding.
However, such 2D absolute spatial encoding alone is insufficient to distinguish between different wavelet levels and sub-band types within each level.
To address this limitation, we extend the standard 2D positional encoding by incorporating a wavelet-level index $j \in \{1,\dots,J\}$ and a sub-band category indicator $d \in \{0,1,2,3\}$, where $d=0$ represents the low-frequency sub-band $\mathbf{L}$, while $d \in \{1,2,3\}$ correspond to distinct high-frequency orientations.
Consequently, as illustrated in \cref{fig4-1}, each token is assigned a 4D positional index $[j,d,\mathrm{Pos}_h,\mathrm{Pos}_w]$, where $\mathrm{Pos}_h$ and $\mathrm{Pos}_w$ denote the spatial coordinates.
The resulting 4D positional index is encoded using the standard sinusoidal frequency-based positional encoding adopted in Vision Transformers~\cite{dosovitskiy2020image}, and subsequently added to the corresponding patch embedding to preserve both spatial layout and wavelet-domain hierarchical information.

Considering the distinct statistical distributions of low- and high-frequency wavelet sub-bands, as well as the presence of residual high-frequency components within the low-frequency sub-band~\cite{unser1997ten, vetterli1995wavelets}, existing approaches that independently refine the two components during global modeling~\cite{huang2024wavedm, miao2024waveface, wang2025wavefusion} may limit the low-frequency representation from fully exploiting the structural cues encoded in the high-frequency sub-bands for effective suppression.
Motivated by this observation, we design a dual-stage Transformer-based interactive refinement module. 
While modeling the low-frequency sub-band and the high-frequency sub-band set separately, we introduce a masked Transformer block to enable cross-frequency interaction, allowing the high-frequency components to guide the suppression of residual high-frequency noise in the low-frequency sub-band. 
The detailed architecture of the $\mathrm{TransBlock}$ is illustrated in \cref{fig3-2}.
  
In the first stage, $\mathbf{S}_L^J$ is fed into a $\mathrm{TransBlock}_L$ without any attention mask to perform preliminary refinement on its smooth structural components. 
The resulting low-frequency tokens are denoted as:
\begin{equation}
\bar{\mathbf{S}}_L^J=\mathrm{TransBlock}_L\!\left(\mathbf{S}_L^J\right).
\label{eq9}
\end{equation}

In the second stage, as illustrated in \cref{fig2}, the input consists of both the high-frequency token set $\mathcal{S}_H=\{\mathbf{S}_H^j\}_{j=1}^{J}$ and the refined low-frequency tokens $\bar{\mathbf{S}}_L^J$. 
To enable cross-frequency interaction, we introduce an attention mask within the $\mathrm{TransBlock}_H$, as illustrated in \cref{fig4-3}. 
Specifically,  high-frequency tokens are visible to low-frequency tokens so that high-frequency structural cues can guide the refinement of the low-frequency component. In contrast, low-frequency tokens  are invisible to high-frequency tokens to avoid reverse interference.
After the second stage, we obtain the further refined low-frequency tokens $\tilde{\mathbf{S}}_L^J$ and high-frequency token set $\tilde{\mathcal{S}}_H$:
\begin{equation}
\tilde{\mathbf{S}}_L^J,
\;
\tilde{\mathcal{S}}_H
=
\mathrm{TransBlock}_H
\left(
\left[
\bar{\mathbf{S}}_L^J,
\mathcal{S}_H
\right],
\mathbf{M}
\right)
\label{eq10}
\end{equation}
where $[\cdot,\cdot]$ denotes concatenation along the token dimension.

For the final transformer refinement outputs, the refined low-frequency sub-band is obtained by summing the two-stage results followed by reshaping:
\begin{equation}
\mathbf{L}_{\mathrm{trans}}^J
=
\mathrm{Reshape}
\left(
\bar{\mathbf{S}}_L^J
+
\tilde{\mathbf{S}}_L^J
\right),
\label{eq11}
\end{equation}
while the refined high-frequency sub-band set is obtained by directly reshaping:
\begin{equation}
\mathcal{H}_{\mathrm{trans}}
=
\mathrm{Reshape}
\left(
\tilde{\mathcal{S}}_H
\right).
\label{eq12}
\end{equation}
\subsubsection{Adaptive Weighted Fusion (AWF).}
\label{sec3-3-3}
Given the complementary modeling capabilities of the LCR and GTR streams, we perform selective adaptive fusion on the wavelet sub-bands generated by the two streams.
Furthermore, considering the distinct statistical distributions of low- and high-frequency wavelet sub-bands, we apply adaptive fusion separately to each component. 
The detailed architecture of the Adaptive Weighted Fusion (AWF) module is illustrated in \cref{fig3-3}.The fused refined wavelet sub-bands are formulated as:
\begin{equation}
\mathbf{L}_{\mathrm{ref}}^{J}
=
\mathrm{AWF}_L
\left(
\mathbf{L}_{\mathrm{conv}}^{J},
\mathbf{L}_{\mathrm{trans}}^{J}
\right),
\label{eq13}
\end{equation}
\vskip -0.1in
\begin{equation}
\mathcal{H}_{\mathrm{ref}}^{j}
=
\mathrm{AWF}_H^{j}
\left(
\mathcal{H}_{\mathrm{conv}}^{j},
\mathcal{H}_{\mathrm{trans}}^{j}
\right),
\quad j \in \{1,\dots,J\},
\label{eq14}
\end{equation}
\subsection{Gaussian-Filtered Frequency Loss (GFFL)}
\label{sec3-4}
In the early sections, we employ $\mathrm{MDWT}$ to explicitly decouple the intermediate representations of both teacher and student networks into low-frequency and high-frequency components.
Specifically, we obtain the teacher sub-bands $\mathbf{L}^{J,t}$ and $\mathcal{H}^{t}$, as well as the student sub-bands.
Considering the representational capacity gap between teacher and student networks, the student wavelet sub-bands typically exhibit greater structural instability. Therefore, we further refine the student sub-bands using the proposed $\mathrm{DS^2SR}$, yielding the refined low-frequency sub-band $\mathbf{L}_{\mathrm{ref}}^{J,s}$ and high-frequency sub-band set $\mathcal{H}_{\mathrm{ref}}^{s}$.

Although wavelet transform effectively captures spatial locality, it does not fully exploit global frequency-domain characteristics.
Moreover, due to intrinsic inductive bias discrepancies across heterogeneous architectures, not all spatial information in the wavelet sub-bands is equally transferable. 
To better leverage frequency-domain properties and selectively emphasize informative components, we propose a Gaussian-Filtered Frequency Loss (GFFL), enabling spatial--frequency joint-aware distillation.

Specifically, for both the teacher wavelet sub-bands $\mathbf{L}^{J,t}, \mathcal{H}^{t}$ and the refined student sub-bands $\mathbf{L}_{\mathrm{ref}}^{J,s}, \mathcal{H}_{\mathrm{ref}}^{s}$, we first apply the Fast Fourier Transform (FFT). 
FFT is an efficient algorithm for computing the discrete Fourier transform (DFT)~\cite{cooley1965algorithm}, reducing computational complexity from $\mathcal{O}(N^2)$ to $\mathcal{O}(N \log N)$. 
The 2D Fourier transform of a sub-band feature $\mathbf{X}$ is defined as:
\begin{equation}
\mathcal{F}(\mathbf{X})(u,v)
=
\sum_{x=0}^{H-1}
\sum_{y=0}^{W-1}
\mathbf{X}(x,y)
\exp
\left(
- j 2\pi
\left(
\frac{ux}{H}
+
\frac{vy}{W}
\right)
\right),
\label{eq15}
\end{equation}
where $(u,v)$ denote frequency coordinates.

As localized spatial structures are explicitly modeled through wavelet decomposition, the frequency-domain component is accordingly designed to emphasize global structural information and spectral energy distribution.
Considering that the amplitude spectrum of Fourier representations effectively encodes global structural information and energy distribution~\cite{jiang2021focal,oppenheim2005importance}, we extract the amplitude spectrum from the Fourier-transformed features. 
Given a feature map $\mathbf{F}$, its amplitude spectrum is computed as:
\begin{equation}
\left|
\mathcal{F}(\mathbf{F})
\right|
=
\sqrt{
\operatorname{Re}\!\left(\mathcal{F}(\mathbf{F})\right)^{2}
+
\operatorname{Im}\!\left(\mathcal{F}(\mathbf{F})\right)^{2}
},
\end{equation}
where $\operatorname{Re}(\cdot)$ and $\operatorname{Im}(\cdot)$ denote the real and imaginary components, respectively.

As different wavelet sub-bands focus on distinct frequency ranges, we employ Gaussian filters to selectively emphasize the dominant frequency components of each sub-band.
Compared with ideal truncation-based filters, Gaussian filters provide smooth transitions in the frequency domain, thereby mitigating boundary artifacts and ringing effects caused by abrupt spectral truncation.
The Gaussian low-pass and high-pass filters are defined as:
\begin{equation}
G_{\mathrm{low}}(u,v)
=
\exp
\left(
-
\frac{u^2+v^2}{2\sigma^2}
\right),
\qquad
G_{\mathrm{high}}(u,v)
=
1 - G_{\mathrm{low}}(u,v).
\label{eq17}
\end{equation}
where $\sigma$ denotes a bandwidth parameter that flexibly controls the emphasis on low-frequency or high-frequency components.

The specific Gaussian filtering masks are illustrated in \cref{fig4-2}. 
For the low-frequency sub-band, the dominant information is primarily concentrated in the low-frequency region (i.e., the central area of the spectrum after FFT and frequency shifting). 
Therefore, a Gaussian low-pass filtering mask is applied.
In contrast, for the high-frequency sub-band set, the dominant components are mainly distributed in the high-frequency regions (i.e., the peripheral area of the shifted spectrum). 
Accordingly, a Gaussian high-pass filtering mask is used.

Given a student sub-band $\mathbf{X}^{s}$ and its corresponding teacher sub-band $\mathbf{X}^{t}$ (either low-frequency or high-frequency), we first compute their Fourier amplitude spectra and apply Gaussian filtering, followed by an InfoNCE loss to align the filtered spectral representations.
The unified formulation of the Gaussian-Filtered Frequency Loss (GFFL) is given by:
\begin{equation}
\mathcal{L}_{\mathrm{GFFL}}
\left(
\mathbf{X}^{s}, \mathbf{X}^{t};\, G
\right)
=
\mathcal{L}_{\mathrm{InfoNCE}}
\left(
G \odot \left| \mathcal{F}(\mathbf{X}^{s}) \right|,
G \odot \left| \mathcal{F}(\mathbf{X}^{t}) \right|
\right),
\label{eq18}
\end{equation}

Inspired by FCFD~\cite{liu2023function}, to enhance task-oriented alignment, we further reconstruct the refined student wavelet sub-bands via IMDWT to obtain a recovered student representation $\mathbf{F}_{\mathrm{rec}}^{s}$.
The $\mathbf{F}_{\mathrm{rec}}^{s}$ is then aligned with the original teacher representation $\mathbf{F}^{t}$ using a GFFL without frequency filtering.
In addition, the $\mathbf{F}_{\mathrm{rec}}^{s}$ is fed into a newly initialized classifier to produce logits $\mathbf{Z}_{\mathrm{rec}}^{s}$, which are supervised by the teacher logits through a Kullback–Leibler (KL) 
divergence loss.
The overall training objective is formulated as:
\begin{equation}
\begin{aligned}
\mathcal{L}_{\mathrm{total}}
&=
\lambda_{1}\,
\mathcal{L}_{\mathrm{GFFL}}
\!\left(
\mathbf{L}_{\mathrm{ref}}^{J,s},
\mathbf{L}^{J,t};
G_{\mathrm{low}}
\right)
+
\lambda_{2}\,
\mathcal{L}_{\mathrm{GFFL}}
\!\left(
\mathcal{H}_{\mathrm{ref}}^{s},
\mathcal{H}^{t};
G_{\mathrm{high}}
\right) \\
&\quad +
\lambda_{3}\,
\mathcal{L}_{\mathrm{GFFL}}
\!\left(
\mathbf{F}_{\mathrm{rec}}^{s},
\mathbf{F}^{t}
\right)
+
\lambda_{4}\,
\mathcal{L}_{\mathrm{KL}}
\!\left(
\mathbf{Z}_{\mathrm{rec}}^{s},
\mathbf{Z}^{t}
\right).
\end{aligned}
\label{eq19}
\end{equation}
\begin{table*}[t]
\renewcommand{\arraystretch}{1.3}
\centering
\caption{\textbf{Comparison of Heterogeneous Distillation Results on CIFAR-100.}  The best and second-best results are emphasized in \textbf{bold} and \underline{underlined} cases.}
\scalebox{0.62}{
\begin{tabular}{cc|cc|cccc|cccccc}
\hline\hline
\multirow{2}{*}{Teacher} & \multirow{2}{*}{Student} &\multicolumn{2}{c|}{From Scratch} & \multicolumn{4}{c|}{Logits-based} & \multicolumn{6}{c}{Feature-based}\\
\cline{3-14} 
        &       & T. & S. & KD~\cite{hinton2015distilling}  & DKD~\cite{zhao2022decoupled} & DIST~\cite{huang2022knowledge} & OFA~\cite{hao2023one} & FitNet~\cite{romero2014fitnets} & CC~\cite{peng2019correlation} & RKD~\cite{park2019relational} & CRD~\cite{tian2019contrastive} & FBT~\cite{li2025fuse} &\textbf{SFKD}\\
\hline
Swin-T & ResNet18 & 89.26 & 74.01 & 78.74 & 80.26 & 77.75 & 80.54 & 78.87 & 74.19 &  74.11 & 77.63 & \underline{81.61} &\cellcolor{gray!20}\textbf{83.26} \\
ViT-S & ResNet18 & 92.04 & 74.01 & 77.26 & 78.10 & 76.49 & 80.15 & 77.71 & 74.26 &  73.72 & 76.60 & \underline{81.93} & \cellcolor{gray!20}\textbf{82.10} \\
Mixer-B/16 & ResNet18 & 87.29 & 74.01 & 77.79 & 78.67 & 76.36 & 79.39 & 77.15 & 74.26 & 73.75 & 76.42 & \underline{81.90} & \cellcolor{gray!20}\textbf{81.98} \\
Swin-T & MobileNetV2 & 89.26 & 73.68 & 74.68 & 71.07 & 72.89 & 80.98 & 74.28 & 71.19 &  69.00 & 79.80 & \underline{81.28} & \cellcolor{gray!20}\textbf{82.03} \\
ViT-S & MobileNetV2 & 92.04 & 73.68 & 72.77 & 69.80 & 72.54 & 78.45 & 73.54 & 70.67 & 68.46 & 78.14 & \textbf{82.10} & \cellcolor{gray!20}\underline{79.97} \\
Mixer-B/16 & MobileNetV2 & 87.29 & 73.68 & 73.33 & 70.20 & 73.26 & 78.78 & 73.78 & 70.73 & 68.95 & 78.15 & \underline{80.83} & \cellcolor{gray!20}\textbf{81.17} \\
\hline
ConvNeXt-T &  DeiT-T  & 88.41 & 68.00 & 72.99 & 74.60 & 73.55 & 75.76 & 60.78 & 68.01 & 69.79 & 65.94 & \underline{79.57} & \cellcolor{gray!20}\textbf{83.91} \\
Mixer-B/16 &  DeiT-T  & 87.29 & 68.00 & 71.36 & 73.44 & 71.67 & 73.90 & 71.05 & 68.13 & 69.89 & 65.35 & \underline{74.40} & \cellcolor{gray!20}\textbf{81.14} \\
ConvNeXt-T &  Swin-P  & 88.41 & 72.63 & 76.44 & 76.80 & 76.41 & 78.32 & 24.06 & 72.63 & 71.73 & 67.09 & \underline{80.73} & \cellcolor{gray!20}\textbf{82.16} \\
Mixer-B/16 &  Swin-P  & 87.29 & 72.63 & 75.93 & 76.39 & 75.85 & 76.65 & 75.20 & 73.32 & 70.82 & 67.03 & \underline{78.44} & \cellcolor{gray!20}\textbf{81.60} \\
\hline
ConvNeXt-T &  ResMLP-S12  & 88.41 & 66.56 & 72.25 & 73.22 & 71.93 & 75.21 & 45.47 & 67.70 & 65.82 & 63.35  & \underline{78.03} &  \cellcolor{gray!20}\textbf{85.08} \\
Swin-T &  ResMLP-S12  & 87.29 & 66.56 & 71.89 & 72.82 & 11.05 & 73.58 & 63.12 & 68.37 & 64.66 & 61.72 & \underline{77.20} & \cellcolor{gray!20}\textbf{84.01} \\
\hline
\multicolumn{2}{c|}{Average Improvements} &  \multicolumn{2}{c|}{}  & $\uparrow$3.12 & $\uparrow$3.16 & $\downarrow$2.31 & $\uparrow$6.19 & $\downarrow$5.21 & $\downarrow$0.33 & $\downarrow$1.39 & $\downarrow$0.02 & \underline{$\uparrow$8.38}  & \cellcolor{gray!20}\textbf{$\uparrow$10.92}\\
\hline\hline
\end{tabular}
}
\vskip -0.2in
\label{table1}
\end{table*}
\vspace{-15pt}
\section{Experiments}
\label{sec4}
\subsection{Implementation Details}
\label{sec4-1}
Detailed descriptions of the \textbf{Models}, \textbf{Datasets}, \textbf{Baselines}, and \textbf{Training Protocols} used in our experiments are provided in Appendix (see \cref{appendix2}).
\subsection{Main Results}
\label{sec4-2}
We conduct extensive experiments on a wide range of teacher–student pairs.
Compared with existing methods, our approach achieves state-of-the-art or competitive performance on both CIFAR-100 and ImageNet-1K.
Across heterogeneous teacher–student pairs, SFKD improves the average Top-1 accuracy by 10.92\% on CIFAR-100 and 2.51\% on ImageNet-1K.

\textbf{Results on CIFAR-100.}
To evaluate the effectiveness of SFKD on CIFAR-100, we conduct experiments on 12 heterogeneous teacher–student pairs.
As shown in \cref{table1}, for feature-based methods, except for FBT~\cite{li2025fuse}, which is specifically designed for heterogeneous distillation, most methods fail to achieve consistent improvements in heterogeneous models, and in some cases even degrade student performance.
This observation aligns with our analysis in \cref{sec1}, where significant representational discrepancies between heterogeneous models make direct feature alignment challenging.
In particular, when the student model is Transformer-based, methods that perform feature projection followed by pixel-wise MSE alignment (\eg, FitNet~\cite{romero2014fitnets}) tend to exhibit highly degraded performance.
For example, in the ConvNeXt-T $\rightarrow$ Swin-P setting, FitNet achieves only 24.06\% accuracy.
For logit-based methods originally designed for homogeneous distillation, the performance gains in heterogeneous settings are generally limited.
FBT~\cite{li2025fuse} and OFA~\cite{hao2023one}, which are specifically designed for heterogeneous distillation, mitigate discrepancies by compressing representations into a compact embedding space or mapping intermediate features to the logit space, respectively.
In contrast, SFKD explicitly decouples and selectively emphasizes spatial information through spatial–frequency joint-aware modeling. 
Compared with the strongest baseline FBT, our method achieves an additional average improvement of 2.54\% on CIFAR-100.

\textbf{Results on ImageNet-1K.}
To further validate the generalization capability of SFKD, we conduct experiments on the larger-scale ImageNet-1K dataset using 14 heterogeneous teacher–student pairs, as shown in \cref{table2}.
Compared with CIFAR-100, ImageNet-1K provides substantially more training data.
As discussed in the ViT~\cite{dosovitskiy2020image}, Transformers typically require larger-scale training to compensate for weaker convolutional inductive biases, and thus benefit more from increased data volume.
Consequently, feature-based distillation methods tend to exhibit more stable improvements on ImageNet-1K than on CIFAR-100, especially when the student is Transformer-based.
For example, in the ResNet-50 $\rightarrow$ DeiT-T setting, FitNet~\cite{romero2014fitnets} improves the DeiT-T accuracy to 75.84\%
Nevertheless, their performance gains remain limited compared with approaches specifically designed for heterogeneous distillation, such as FBT~\cite{li2025fuse} and OFA~\cite{hao2023one}. 
Moreover, their performance can be unstable and may even degrade the student model in certain cases. 
For instance, in the ConvNeXt-T $\rightarrow$ DeiT-T pair, FitNet results in lower accuracy than the baseline without distillation.
On ImageNet-1K, SFKD consistently achieves the best or competitive performance across heterogeneous settings, further demonstrating the effectiveness and generality of explicitly decoupling and selectively exploiting spatial information in heterogeneous representations. 
Moreover, compared with OFA, which utilizes features from four stages for distillation, our method relies solely on the final representation, similar to FBT, while still achieving comparable or superior performance.
\begin{table*}[t]
\renewcommand{\arraystretch}{1.3}
\centering
\caption{\textbf{Comparison of Heterogeneous Distillation Results on ImageNet-1K.} The best and second-best results are emphasized in \textbf{bold} and \underline{underlined} cases.}
\scalebox{0.62}{
\begin{tabular}{cc|cc|cccc|cccccc}
\hline\hline
\multirow{2}{*}{Teacher} & \multirow{2}{*}{Student} &\multicolumn{2}{c|}{From Scratch} & \multicolumn{4}{c|}{Logits-based} & \multicolumn{6}{c}{Feature-based}\\
\cline{3-14} 
        &       & T. & S. & KD~\cite{hinton2015distilling}  & DKD~\cite{zhao2022decoupled} & DIST~\cite{huang2022knowledge} & OFA~\cite{hao2023one} & FitNet~\cite{romero2014fitnets} & CC~\cite{peng2019correlation} & RKD~\cite{park2019relational} & CRD~\cite{tian2019contrastive} & FBT~\cite{li2025fuse} &\textbf{SFKD}\\
\hline
Deit-T & ResNet18 & 72.17 & 69.75 & 70.22 & 69.39 & 70.64 & 71.01 & 70.44 & 69.77 & 69.47 & 69.25 & \underline{71.22} & \cellcolor{gray!20}\textbf{71.24}\\
Swin-T & ResNet18 & 81.38 & 69.75 & 71.14 & 71.10 & 70.91 & 71.76 & 71.18 & 70.07 & 68.89 & 69.09 & \underline{72.21} & \cellcolor{gray!20}\textbf{72.54}\\
Mixer-B/16 & ResNet18 & 76.62 & 69.75 & 70.89 & 69.89 & 70.66 & 71.38 & 70.78 & 70.05 & 69.46 & 68.40 & \underline{71.44} & \cellcolor{gray!20}\textbf{71.71}\\
Deit-T & MobileNetV2 & 72.17 & 68.87 & 70.87 & 70.14 & 71.08 & 71.39 & 70.95 & 70.69 & 69.72 & 69.60 & \underline{71.78} & \cellcolor{gray!20}\textbf{71.83}\\
Swin-T & MobileNetV2 & 81.38 & 68.87 & 72.05 & 71.71 & 71.76 & 72.32 & 71.75 & 70.69 & 67.52 & 69.58 & \underline{72.54} & \cellcolor{gray!20}\textbf{72.67}\\
Mixer-B/16 & MobileNetV2 & 76.62 & 68.87 & 71.92 & 70.93 & 71.74 & 72.12 & 71.59 & 70.79 & 69.86 & 68.89 & \underline{72.31} & \cellcolor{gray!20}\textbf{72.43}\\
\hline
ResNet50 & Deit-T & 80.38 & 72.17 & 75.10 & 75.60 & 75.13 & 75.73 & \underline{75.84} & 72.56 & 72.06 & 68.53 & 75.64 & \cellcolor{gray!20}\textbf{75.89}\\
ConvNeXt-T & Deit-T & 82.05 & 72.17 & 74.00 & 73.95 & 74.07 & 74.41 & 70.45 & 73.12 & 71.47 & 69.18 & \underline{75.26} & \cellcolor{gray!20}\textbf{75.29}\\
Mixer-B/16 & Deit-T & 76.62 & 72.17 & 74.16 & 72.82 & 74.22 & 74.46 & 74.38 & 72.82 & 72.24 & 68.23 & \textbf{75.00} & \cellcolor{gray!20}\underline{74.69}\\
ResNet50 & Swin-N & 80.38 & 75.53 & 77.58 & 76.24 & 77.29 & 77.76 & 76.83 & 76.05 & 75.90 & 73.90 & \underline{77.79} & \cellcolor{gray!20}\textbf{78.28}\\
ConvNeXt-T & Swin-N  & 82.05 & 75.53 & 77.15 & 77.00 & 77.25 & 77.50 & 74.81 & 75.79 & 75.48 & 74.15 & \underline{77.73} &\cellcolor{gray!20}\textbf{77.83} \\
Mixer-B/16 & Swin-N  & 76.62 & 75.53 & 76.26 & 75.03 & 76.54 & 76.63 & 76.17 & 75.81 & 75.52 & 73.38 & \textbf{76.87} &
\cellcolor{gray!20}\underline{76.67}\\
\hline
ConvNeXt-T &  ResMLP-S12 & 82.05 & 76.65 & 76.87 & 77.23 & 77.24 & 77.26 & 74.69 & 75.79 & 75.28 & 73.57 & \underline{77.33} & \cellcolor{gray!20}\textbf{78.35} \\
Swin-T &  ResMLP-S12 & 81.38 & 76.65 & 76.67 & 76.99 & 77.25 & 77.31 & 76.48 & 76.15 & 75.10 & 73.40 & \underline{77.42} & \cellcolor{gray!20} \textbf{77.88} \\
\hline
\multicolumn{2}{c|}{Average Improvements} &  \multicolumn{2}{c|}{}  & $\uparrow$1.61 & $\uparrow$1.05 & $\uparrow$1.65 & $\uparrow$2.05 & $\uparrow$1.00 & $\uparrow$0.56 & $\downarrow$0.30 & $\downarrow$1.65 & \underline{$\uparrow$2.31}  & \cellcolor{gray!20}\textbf{$\uparrow$2.51}\\
\hline\hline
\end{tabular}
}
\vskip -0.3in
\label{table2}
\end{table*}
\begin{table*}[t]
\renewcommand{\arraystretch}{1.3}
\centering
\caption{\textbf{Comparison of Homogeneous Distillation Results on ImageNet-1K.} The best and second-best results are emphasized in \textbf{bold} and \underline{underlined} cases.}
\scalebox{0.66}{
\begin{tabular}{cc|cc|ccccccccc|c}
\hline\hline
 &  & T. & S. & AT~\cite{zagoruyko2016paying} & OFD~\cite{heo2019comprehensive} & CRD~\cite{tian2019contrastive} & Review~\cite{chen2021distilling} & DKD~\cite{zhao2022decoupled} & DIST~\cite{huang2022knowledge} & FCFD~\cite{liu2023function} & OFA~\cite{hao2023one} & FBT~\cite{li2025fuse} & \textbf{SFKD} \\
\hline
ResNet34 & ResNet18 & 73.31 & 69.75 & 70.69 & 70.81 & 71.17 & 71.61 & 71.70  & 72.07 & 72.24 & 72.10 & \underline{72.29} & \cellcolor{gray!20}\textbf{72.34}\\
\hline
ResNet50 & MobileNet & 76.61 & 68.58 & 69.56 & 71.25 & 71.37 & 72.56 & 72.05  & 73.24 & 73.37 & 73.28 & \underline{73.45} & \cellcolor{gray!20}\textbf{73.81}\\
\hline\hline
\end{tabular}
}
\label{table3}
\end{table*}

\vspace{-12pt}
\textbf{Results on Homogeneous Distillation.}
To further verify the generality and effectiveness of decoupling and selectively exploiting spatial information in representations through spatial–frequency joint awareness in SFKD, we also conduct experiments under homogeneous distillation settings, as shown in \cref{table3}.
We evaluate two commonly used homogeneous teacher–student pairs: ResNet34 $\rightarrow$ ResNet18 and ResNet50 $\rightarrow$ MobileNet.
As shown in the \cref{table3}, SFKD achieves the best performance in both settings, outperforming existing homogeneous and heterogeneous distillation approaches.
These results indicate that even when the representational discrepancy between teacher and student is relatively small, the proposed strategy of decoupling and selectively exploiting spatial information in representations remains effective for improving distillation performance.
\begin{table}[t]
\centering
\caption{\textbf{Ablation studies on loss components and DS$^2$SR module.} The best and second-best results are highlighted in \textbf{bold} and \underline{underlined}, respectively.}
\label{tab:combined_ablation}
\scriptsize
\setlength{\tabcolsep}{6pt}
\renewcommand{\arraystretch}{1.2}
\begin{subtable}[t]{0.45\linewidth}
\centering
\vskip -0.09in
\caption{\textbf{Ablation study on the effectiveness of each loss component.}}
\label{tab:loss_ablation}
\begin{tabular}{cccc|c}
\hline\hline
$\mathcal{L}_{1}^{LF}$ & $\mathcal{L}_{2}^{HF}$ & $\mathcal{L}_{3}^{Rec}$ & $\mathcal{L}_{4}^{KL}$ & Acc \\
\midrule
$\times$ & $\checkmark$ & $\checkmark$ & $\checkmark$ & 72.19 \\
$\checkmark$ & $\times$ & $\checkmark$ & $\checkmark$ & 72.32 \\
$\checkmark$ & $\checkmark$ & $\times$ & $\checkmark$ & 72.31 \\
$\checkmark$ & $\checkmark$ & $\checkmark$ & $\times$ & \underline{72.46} \\
$\checkmark$ & $\checkmark$ & $\checkmark$ & $\checkmark$ & \textbf{72.54} \\
\hline\hline
\end{tabular}
\label{table4_a}
\end{subtable}
\hspace{0.01\linewidth}
\begin{subtable}[t]{0.49\linewidth}
\centering
\vskip -0.2in
\caption{\textbf{Ablation study on the necessity of wavelet sub-band refinement and the effectiveness of the DS$^2$SR module.}}
\label{tab:dssr_ablation}
\begin{tabular}{cccc}
\hline\hline
\multirow{2}{*}{LCR} & \multirow{2}{*}{GTR} & Swin-T & ResNet50 \\
& & ResNet18 & Swin-N \\
\midrule
$\times$ & $\times$ & 71.24 & 76.83 \\
$\checkmark$ & $\times$ & 72.19 & \underline{78.24} \\
$\times$ & $\checkmark$ & \underline{72.46} & 77.83 \\
$\checkmark$ & $\checkmark$ & \textbf{72.54} & \textbf{78.28} \\
\hline\hline
\end{tabular}
\label{table4_b}
\end{subtable}
\vskip -0.2in
\label{table4}
\end{table}
\vspace{-10pt}
\subsection{Ablation Study}
\label{sec4-3}
\subsubsection{Necessity of Wavelet Sub-band Refinement and Effectiveness of the DS$^2$SR Module.}
\label{sec4-3-1}
As discussed in \cref{sec3-3}, wavelet decomposition cannot achieve ideal band-limited separation~\cite{mallat1999wavelet, strang1996wavelets}, which may lead to spectral leakage or residual frequency overlap between sub-bands~\cite{unser1997ten, vetterli1995wavelets}. 
To address this issue, and considering the distinct inductive biases across heterogeneous models, we design a novel and universal DS$^2$SR module to further refine the wavelet sub-bands.
To evaluate the effectiveness of each refinement stream, we conduct experiments on ImageNet-1K using two heterogeneous teacher–student pairs: Swin-T $\rightarrow$ ResNet18 and ResNet50 $\rightarrow$ Swin-N. 
The results are summarized in \cref{table4_b}. 
When no refinement stream is introduced, the distillation performance remains limited for both CNN-based and Transformer-based students. This observation, as discussed in \cref{sec3-1}, further confirms that the weaker representation capability of student models leads to more unstable representations that require further refinement.
After introducing either refinement stream, the performance of both student models improves noticeably.
Further analysis reveals that the CNN student (ResNet18) is more sensitive to the GTR stream, achieving an additional improvement of 0.27\% compared with using only the LCR stream. 
In contrast, the Transformer student (Swin-N) benefits more from the LCR stream, yielding an additional improvement of 0.41\% compared with using only the GTR stream.
We believe this observation is consistent with the analysis in FBT~\cite{li2025fuse}, indicating that introducing streams with complementary inductive biases can better facilitate the student model in capturing knowledge from heterogeneous models.
Finally, by adaptively combining the two streams, the DS$^2$SR module achieves the best performance of both student models.
\vspace{-12pt}
\subsubsection{Effectiveness of Each Loss Components.}
As shown in \cref{eq19}, the proposed objective consists of four loss terms, denoted as $\mathcal{L}_1^{LF}$, $\mathcal{L}_2^{HF}$, $\mathcal{L}_3^{Rec}$, and $\mathcal{L}_4^{KL}$, respectively.
To investigate the contribution of each loss component, we conduct an ablation study on the Swin-T$\rightarrow$ResNet18 teacher-student pair over ImageNet-1K.
The results are summarized in \cref{table4_a}.
The results indicate that removing $\mathcal{L}_1^{LF}$ causes the largest performance degradation, demonstrating that the low-frequency sub-band, which mainly preserves the global structural information of representations, plays critical role in knowledge transfer.
Removing $\mathcal{L}_2^{HF}$ also results in a noticeable performance drop, indicating that the high-frequency sub-band set provides complementary local structural details that are likewise essential for effective distillation.
Both $\mathcal{L}_1^{LF}$ and $\mathcal{L}_2^{HF}$ perform alignment at the wavelet sub-band level. 
Specifically, Gaussian low- and high-pass filters are introduced to selectively emphasize the dominant frequency components of different sub-bands, while the objective in \cref{eq18} is essentially an InfoNCE-style isotropic contrastive objective. 
However, as demonstrated in FCFD~\cite{liu2023function}, neural networks utilize intermediate features in an anisotropic manner.
Consequently, performing only sub-band level alignment with an isotropic objective may still introduce structural discrepancies between $\mathbf{F}_{\mathrm{rec}}^{s}$ and $\mathbf{F}^{t}$.
To alleviate this issue, $\mathcal{L}_3^{Rec}$ is introduced to further constrain the reconstructed feature $\mathbf{F}_{\mathrm{rec}}^{s}$ after $\mathrm{IMDWT}$, encouraging overall structural consistency with $\mathbf{F}^{t}$. 
In addition, $\mathcal{L}_4^{KL}$ provides task-oriented supervision to ensure that the transferred knowledge remains beneficial for the downstream classification objective. As shown in Table~\ref{table4_a}, removing either $\mathcal{L}_3^{Rec}$ or $\mathcal{L}_4^{KL}$ consistently degrades the student performance, further validating the effectiveness of all four loss components.
\vspace{-12pt}
\subsubsection{Extended Experiments.}
We provide additional experiments in Appendix (see \cref{appendix3}) to further evaluate SFKD in a more comprehensive manner.
\vspace{-8pt}
\section{Conclusion and Future Work}
\label{sec5}
We discuss \textbf{Conclusion and Future Work} and defer a concentrated account to Appendix (see \cref{appendix4}).
\vspace{-8pt}
\section{Acknowledgements}
\label{sec6}
This work was supported by the National Natural Science Foundation of China under Grant 62271153.

\bibliographystyle{splncs04}
\bibliography{main}

\clearpage
\setcounter{page}{1}
\title{\scshape SFKD: Spatial--Frequency Joint-Aware Heterogeneous Knowledge Distillation via Multi-Level Wavelet Spectral Interaction\\
\normalfont (Appendix)} 

\titlerunning{SFKD}

\author{Cuipeng Wang\inst{1,2}\orcidlink{0009-0009-6299-4729} \and
Haipeng Wang\inst{1,2}\orcidlink{0000-0003-1912-7143}\thanks{Corresponding author.}}

\authorrunning{C.~Wang et al.}

\institute{
Key Laboratory for Information Science of Electromagnetic Waves,\\
Ministry of Education, Fudan University, Shanghai, China\\
\and
Discipline and Technology Center of Microwave Vision Intelligent Sensing,\\
Fudan University, Shanghai, China\\
\email{cpwang23@m.fudan.edu.cn, hpwang@fudan.edu.cn}
}
\maketitle

\appendix

\section{Related Work}
\label{appendix1}
\subsection{Knowledge distillation}
Knowledge distillation improves the performance of a compact student network by transferring knowledge from a powerful teacher network without increasing the model size.
According to the location where distillation is performed, existing knowledge distillation methods can generally be categorized into logit-based distillation~\cite{hinton2015distilling,huang2022knowledge,zhao2022decoupled,hao2023one} and feature-based distillation~\cite{romero2014fitnets,tian2019contrastive,chen2021distilling,heo2019comprehensive,park2019relational,peng2019correlation,liu2023function,li2025fuse,wu2024aligning,zagoruyko2016paying}.

Most existing knowledge distillation methods~\cite{hinton2015distilling,huang2022knowledge,zhao2022decoupled,romero2014fitnets,tian2019contrastive,chen2021distilling,heo2019comprehensive,park2019relational,peng2019correlation,liu2023function} primarily focus on homogeneous models (\eg, CNN$\rightarrow$CNN distillation). 
With the remarkable success of Transformer and MLP-based models in computer vision, which have achieved performance comparable to CNNs, an increasing number of studies have begun to explore whether knowledge transfer can be achieved across heterogeneous architectures.
Benefiting from prior studies on heterogeneous distillation~\cite{li2025fuse,hao2023one,wu2024aligning} as well as hybrid architectures~\cite{li2023convmlp, li2023uniformer}, it has been observed that transferring knowledge across heterogeneous models with distinct yet complementary inductive biases can often yield unexpected performance gains.

As discussed in \cref{sec1}, distillation restricted to homogeneous architectures suffers from certain limitations.
To broaden the range of candidate teacher models and enable student networks to better exploit cross-architecture knowledge, several studies have investigated how to mitigate representational discrepancies between heterogeneous models.
Hao et al.~\cite{hao2023one} observe substantial representational differences across heterogeneous architectures and propose to map the student features at each stage into the logit space via dedicated projectors, followed by alignment with the teacher logits using a novel OFA loss. 
Wu et al.~\cite{wu2024aligning} introduce multi-scale low-pass filters to compress both teacher and student representations into a compact feature space before performing alignment. 
Li et al.~\cite{li2025fuse} draw inspiration from hybrid architectures and construct a bridging network between teacher and student models to generate intermediate representations, enabling knowledge transfer by applying spatially agnostic losses to the teacher, student, and intermediate representations.
Due to the significant discrepancies in spatial information across heterogeneous representations, existing methods often discard or suppress such spatial information. 
However, spatial information in representations often encodes both transferable global structural semantics and architecture-specific local details. 
Therefore, we argue that spatial information in heterogeneous representations should not be directly ignored, but rather selectively exploited to facilitate more effective knowledge transfer.

To effectively exploit the spatial information in heterogeneous representations, we propose a Spatial--Frequency Joint-Aware Heterogeneous Knowledge Distillation framework (\textbf{SFKD}). 
By explicitly decoupling spatial information in heterogeneous representations and selectively enhancing informative components, our approach better uncovers meaningful knowledge across heterogeneous architectures, enabling more efficient knowledge transfer.
\subsection{Discrete Wavelet Transform}
In SFKD, the explicit decoupling of spatial information in representations is achieved through multi-level discrete wavelet transform (MDWT). 
DWT is a widely used signal processing technique that decomposes an input feature into a low-frequency (LF) sub-band and a set of high-frequency (HF) sub-bands, thereby facilitating explicit spatial decoupling.
Wavelet transforms have attracted increasing attention in computer vision and have been successfully incorporated into deep learning models to improve efficiency and performance across a variety of visual tasks.

As wavelet transforms can decouple representations while preserving informative structures, several studies have adopted them to replace conventional pooling or downsampling operations, thereby improving information preservation and computational efficiency.
For example, Finder et al.~\cite{finder2024wavelet} integrate wavelet transforms with convolution operations, enabling a backbone architecture that achieves a larger receptive field with smaller convolution kernels. 
Yao et al.~\cite{yao2022wave} introduce wavelet transforms into Transformer architectures by performing downsampling of key–value tokens via discrete wavelet transform, thereby reducing computational complexity, while inverse wavelet transforms are used to restore global contextual information. 
In addition, wavelet transforms have demonstrated effectiveness in various vision tasks such as image super-resolution~\cite{du2025diffusion} and image denoising~\cite{li2024ewt}.

Motivated by the ability of wavelet transforms to explicitly decouple spatial information, we explore leveraging such decomposition to decouple heterogeneous representations and selectively exploit informative components for more effective knowledge transfer.
To the best of our knowledge, the use of wavelet transforms in heterogeneous knowledge distillation remains largely unexplored.
In this work, we are the first to introduce multi-level wavelet transforms into heterogeneous distillation to explicitly decouple spatial information in intermediate representations. 
Furthermore, we design a dual-stream dual-stage spectral refinement (DS$^2$SR) module to further refine the wavelet sub-bands of student representations, enabling more effective alignment with the teacher representations.
\subsection{Fast Fourier Transform}
Although wavelet transforms effectively capture localized spatial structures, they do not fully exploit frequency-domain characteristics.
Moreover, we observe that heterogeneous models often exhibit significant discrepancies in the spatial information encoded in their representations, which makes direct spatial-domain alignment challenging.
Prior studies have shown that Fourier frequency representations are more effective at capturing global information in representations.
Therefore, we explore transferring the alignment of heterogeneous representations from the spatial domain to the frequency domain.

In Fourier representations, the amplitude spectrum effectively encodes global structural information and energy distribution, while the phase spectrum mainly characterizes local structural details~\cite{oppenheim2005importance}.
Fast Fourier transform has been widely applied across various computer vision tasks~\cite{jiang2021focal}, and several studies have explored integrating Fourier representations with knowledge distillation.

In this work, we combine wavelet transforms with Fast Fourier Transform for heterogeneous knowledge distillation. 
Specifically, we first apply multi-level wavelet transforms to explicitly decouple spatial information in heterogeneous representations. 
The resulting wavelet sub-bands are then transformed into the Fourier domain, where a novel Gaussian-filtered frequency loss is designed to selectively emphasize informative frequency components in different sub-bands. 
This combination enables a spatial–frequency joint-aware distillation framework.
\begin{table*}[t]
\centering
\setlength{\tabcolsep}{5pt}
\renewcommand{\arraystretch}{1.2}
\caption{\textbf{Training hyperparameters on CIFAR100 and ImageNet-1K.}
Our training hyperparameters are consistent with those in OFA~\cite{hao2023one}, with differences from the FBT~\cite{li2025fuse} settings highlighted in \textbf{bold}, and the FBT parameter settings are marked in \textcolor{red}{red} within parentheses.}
\begin{tabular}{l|cc|cc}
\hline\hline
   & \multicolumn{2}{c|}{ImageNet-1K} & \multicolumn{2}{c}{CIFAR100} \\
\cline{2-5}
 & CNN & MSA/MLP & CNN & MSA/MLP \\
\hline
Epochs & 100 & 300 & 300 & 300 \\
Image resolution & $224^2$ & $224^2$ & $224^2$ & $224^2$ \\
Batch size & \textbf{64\textcolor{red}{\ (512)}} & \textbf{128 \textcolor{red}{\ (1024)}} & \textbf{128 \textcolor{red}{\ (1024)}} & \textbf{128 \textcolor{red}{\ (512)}} \\
Initial LR & 0.1 & $5e^{-4}$ & \textbf{0.05 \textcolor{red}{\ (0.1)}} & $5e^{-4}$ \\
Minimum LR & $1e^{-6}$ & $1e^{-6}$ & $1e^{-3}$ & $1e^{-5}$ \\
Optimizer & SGD & AdamW & SGD & AdamW \\
Weight decay & $1e^{-4}$ & $5e^{-2}$ & $2e^{-3}$ & $5e^{-2}$ \\
LR schedule & $\times 0.1$ at [30,60,90] & Cosine & Cosine & Cosine \\
Warmup & 3 & 20 & 3 & 20 \\
EMA & - & 0.99996 & - & - \\
RandAugment & - & 9/0.5 & - & 9/0.5 \\
Mixup & - & 0.8 & - & 0.8 \\
Cutmix & - & 1.0 & - & 1.0 \\
RE prob & - & 0.25 & - & 0.25 \\
\hline\hline
\end{tabular}
\label{tabapp1}
\vskip -0.2in
\end{table*}
\section{Implementation Details}
\label{appendix2}
\subsection{Models.}
We follow the teacher–student pair settings in OFA~\cite{hao2023one} and conduct both heterogeneous and homogeneous distillation experiments.
The evaluated models span CNN, Transformer, and MLP architectures.
Specifically, the CNN models include ResNet~\cite{he2016deep}, MobileNet~\cite{howard2017mobilenets}, MobileNetV2~\cite{sandler2018mobilenetv2}, and ConvNeXt~\cite{liu2022convnet}.
The Transformer models include ViT~\cite{dosovitskiy2020image}, DeiT~\cite{touvron2021training}, and Swin~\cite{liu2021swin}, while the MLP models consist of MLP-Mixer~\cite{tolstikhin2021mlp} and ResMLP~\cite{touvron2022resmlp}.
\subsubsection{Datasets.}
Experiments are conducted on CIFAR-100 and ImageNet-1K.
CIFAR-100~\cite{krizhevsky2009learning} contains 100 categories with 50,000 training images and 10,000 test images, each of size $32 \times 32$. 
ImageNet-1K~\cite{russakovsky2015imagenet} is a large-scale image classification dataset with 1,000 categories, containing approximately 1.28M training images and 50,000 validation images.
\subsection{Baselines.}
Following FBT~\cite{li2025fuse}, we compare our method with representative and publicly available knowledge distillation approaches.
For logit-based methods, we include KD~\cite{hinton2015distilling}, DKD~\cite{zhao2022decoupled}, DIST~\cite{huang2022knowledge}, and OFA~\cite{hao2023one}.
For feature-based methods, we compare against FitNet~\cite{romero2014fitnets}, ReviewKD~\cite{chen2021distilling}, CC~\cite{peng2019correlation}, RKD~\cite{park2019relational}, CRD~\cite{tian2019contrastive}, FCFD~\cite{liu2023function}, AT~\cite{zagoruyko2016paying}, OFD~\cite{heo2019comprehensive} and FBT~\cite{li2025fuse}.
The baseline results are adopted from those reported in FBT.
\subsection{Training Protocols.}
In this work we follow the training protocols of OFA~\cite{hao2023one}.
However, the experimental settings used in FBT~\cite{li2025fuse} differ from those adopted in both our method and OFA.
Specifically, FBT uses a batch size that is 4–8× larger than ours. 
We conduct experiments on both ImageNet-1K and CIFAR-100.
For CIFAR-100, all models are trained for 300 epochs. 
Specifically, CNN student models are optimized using SGD optimizer, while students with Transformer or MLP architectures are trained using AdamW.
For ImageNet-1K, CNN student models are trained for 100 epochs using SGD optimizer, whereas Transformer or MLP student models are trained for 300 epochs with the AdamW optimizer.
The experiments on CIFAR-100 are conducted on a single NVIDIA 3090 GPU, while the experiments on ImageNet-1K are conducted on four NVIDIA 3090 GPUs.
All experiments are conducted three times, and the results are reported as the average value.
More detailed configurations are summarized in \cref{tabapp1}.
\section{Extended Experiments}
\label{appendix3}
\subsection{Comparison Experiment on the Dual-Stage Module in the LCR Stream}
To evaluate the efficiency of the local convolutional refinement (LCR) stream design, we conduct experiments using the ResNet50$\rightarrow$Swin-N teacher-student pair.
To ensure experimental rigor, the GTR stream is not introduced in this case.
Specifically, we first perform an experiment without ResConvBlock refinement, then another where we apply a single ResConvBlock for refinement without distinguishing between low-frequency and high-frequency sub-bands.
We then apply the designed LCR stream, where the low- and high-frequency sub-bands are fed into two separate ResConvBlock modules for refinement. 
The detailed results are shown in~\cref{tabapp2}.

As analyzed in~\cref{sec3-1}, the results indicate that, due to the weaker representation capability of the student network compared to the teacher network, when the student wavelet sub-bands are not refined, effective alignment with the teacher sub-bands is hindered.
Furthermore, when low- and high-frequency sub-bands are not processed separately, the performance tends to be suboptimal (78.00\%) due to the differing distributions between the low- and high-frequency sub-bands.
In contrast, when the LCR stream is employed, the best performance (78.24\%) is achieved.

\begin{table*}[ht]
\vskip -0.2in
\centering
\setlength{\tabcolsep}{15pt}
\renewcommand{\arraystretch}{1.3}
\caption{\textbf{Comparison experiments of different structural convolutional refinement streams on ImageNet-1K.} The best results are emphasized in \textbf{bold} cases.}
\begin{tabular}{c|c}
\hline\hline
   &  ResNet50 \\
    & Swin-N \\
\hline
No ResConvBlock      & 76.83 \\ 
Single ResConvBlock  & 78.00 \\ 
LCR Stream           & \textbf{78.24} \\ 
\hline\hline
\end{tabular}
\label{tabapp2}
\vskip -0.3in
\end{table*}

\subsection{Comparison Experiment on the Dual-Stage Module in the GTR Stream}
In the design of the global transformer refinement (GTR) stream, we take advantage of the transformer module's ability to model long-range dependencies, thereby enhancing global structural consistency.
To better leverage the characteristics of the transformer module, we design a dual-stage refinement module that facilitates interaction between low- and high-frequency sub-bands. 
Considering the distinct statistical distributions of low- and high-frequency wavelet sub-bands, as well as the presence of residual high-frequency components within the low-frequency sub-band~\cite{unser1997ten, vetterli1995wavelets}, we refine the low- and high-frequency sub-bands separately. 
Meanwhile, while refining these sub-bands, we simultaneously use the GTR stream to leverage the structural information from the high-frequency sub-bands to suppress the residual high-frequency components in the low-frequency sub-bands.
To ensure experimental rigor, the LCR stream is not introduced in this case.
Compared to the GTR stream, the LCR stream design is more refined. To more effectively demonstrate the efficiency of this design, we present additional comparative experiments.
In the loss function section, we use both the commonly used MSE loss and our proposed GFFL loss function to better validate its generalization capability. 
We conduct experiments using the Swin-T$\rightarrow$ResNet18 teacher-student pair.
The detailed results are shown in~\cref{tabapp3}.

The results show that, regardless of the loss function used, the performance is suboptimal when wavelet sub-bands are not refined.
In this experiment, we use both a single TransBlock and two TransBlocks concatenated directly, without introducing any masks and without distinguishing between low- and high-frequency wavelet sub-bands.
We find that using two TransBlocks in a concatenated manner leads to relatively better performance, but the improvement remains limited. 
Due to the differing distributions of low- and high-frequency wavelet sub-bands, we refine them separately and introduce a mask to guide the low-frequency sub-bands in removing residual high-frequency components using the high-frequency sub-bands. 
Ultimately, we achieve the best performance in both loss functions.

\begin{table*}[ht]
\vskip -0.2in
\centering
\setlength{\tabcolsep}{12pt}
\renewcommand{\arraystretch}{1.3}
\caption{\textbf{Comparison experiments of different structural transformer refinement streams on ImageNet-1K.} The first column uses MSE loss, and the second column uses GFFL loss. The best results are emphasized in \textbf{bold} cases.}
\begin{tabular}{c|c|c}
\hline\hline
   &  Swin-T &  Swin-T\\
    & ResNet18/MSE & ResNet18/GFFL\\
\hline
No TransBlock                     & 71.65  & 71.24\\ 
Single TransBlock                 & 71.70  & 72.10\\ 
Two TransBlocks Cascaded          & 71.86 & 72.31\\ 
GTR Stream                        & \textbf{71.95} & \textbf{72.46}\\ 
\hline\hline
\end{tabular}
\label{tabapp3}
\vskip -0.3in
\end{table*}

\subsection{Comparison Experiment of the Gaussian-Filtered Frequency Loss Function}
As analyzed in~\cref{sec3-4}, we observe that significant spatial domain discrepancies exist between heterogeneous representations, making direct alignment in the spatial domain challenging.
As shown in~\cref{fig1}, after applying Fast Fourier Transform (FFT) to the wavelet sub-bands, the differences in their frequency-domain amplitude spectra are relatively smaller.
This is attributable to the amplitude spectrum captures global structural information and energy distribution~\cite{oppenheim2005importance}.
Based on this, we design a Gaussian-filtered frequency loss function (GFFL). To evaluate its effectiveness, we conduct a comparative experiment using the Mixer-B/16$\rightarrow$Swin-N teacher-student pair, with the results presented in~\cref{tabapp4}.

We compare the Gaussian-filtered frequency loss (GFFL) with commonly used MSE loss and InfoNCE loss functions. 
As discussed in~\cref{sec1}, directly discarding or suppressing spatial information often impedes the transfer of useful information, resulting in suboptimal performance. 
Compared to MSE loss, the use of InfoNCE loss compresses the spatial information in representations, leading to the poorest performance (75.73\%). In contrast, MSE loss, calculated pixel-wise in the spatial domain, retains some spatial information, yielding relatively better performance (76.10\%). 
Leveraging the Fourier domain's ability to capture global structural information and energy distribution, as well as the different frequency components emphasized in each wavelet sub-band, we apply Gaussian filtering to selectively emphasize informative components in each sub-band, ultimately achieving the best performance of 76.67\%.

\begin{table*}[t]
\centering
\setlength{\tabcolsep}{20pt}
\renewcommand{\arraystretch}{1.3}
\caption{\textbf{Comparison of Different Loss Functions on ImageNet-1K.} }
\begin{tabular}{c|c}
\hline\hline
\multirow{2}{*}{Loss Function Type}    &  Mixer-B/16 \\
                                       & Swin-N \\
\hline
MSE Loss                               & 76.10 \\ 
InfoNCE Loss                           & 75.73 \\ 
GFFL                                   & \textbf{76.67} \\ 
\hline\hline
\end{tabular}
\label{tabapp4}
\vskip -0.2in
\end{table*}
\begin{table*}[ht]
\vspace{-0.2in}
\centering
\caption{\textbf{Comparison of Accuracy and Computational Cost on ImageNet-1K.} The best and second-best results are highlighted in \textbf{bold} and \underline{underlined}, respectively.}
\renewcommand{\arraystretch}{1.3}
\resizebox{\textwidth}{!}{%
\begin{tabular}{l l l c c c c c c}
\toprule
\textbf{T.} & \textbf{S.} & \textbf{Method} 
& \textbf{Acc. $\uparrow$}
& \textbf{Extra Params (M) $\downarrow$}
& \textbf{Train FLOPs (G) $\downarrow$}
& \textbf{Extra FLOPs (G) $\downarrow$}
& \textbf{Memory (GB) $\downarrow$}
& \textbf{Runtime (s/epoch) $\downarrow$} \\
\midrule

Swin-T & ResNet18 & FitNet 
& 71.18 
& \underline{11.02}
& 13.32 
& 3.48 
& \underline{8.9}
& 1312.24 \\

Swin-T & ResNet18 & OFA 
& \underline{71.76} 
& \textbf{4.39} 
& \textbf{10.12} 
& \textbf{0.28}
& \textbf{8.7}
& \textbf{1207.91} \\

Swin-T & ResNet18 & SFKD 
& \textbf{72.54} 
& 29.80
& \underline{12.97}
& \underline{3.13} 
& \underline{8.9}
& \underline{1280.90} \\

\midrule

DeiT-T & MobileNetV2 & FitNet 
& 70.95 
& \textbf{1.09}
& \underline{2.57}
& \underline{0.58} 
& 12.1
& 1180.33 \\

DeiT-T & MobileNetV2 & OFA 
& \underline{71.39} 
& 7.14 
& \textbf{2.31}
& \textbf{0.32}
& \textbf{11.6}
& \textbf{1070.74} \\

DeiT-T & MobileNetV2 & SFKD 
& \textbf{71.83} 
& \underline{2.24}
& 2.61
& 0.62
& \underline{11.9} 
& \underline{1074.31} \\

\midrule

ConvNeXt-T & Swin-N & FitNet 
& 74.81 
& \textbf{11.02}
& \underline{11.94}
& \underline{3.48}
& 23.54
& 1162.92 \\

ConvNeXt-T & Swin-N & OFA 
& \underline{77.50} 
& \underline{28.19} 
& 13.19
& 4.73
& \underline{22.55} 
& \underline{1427.62} \\

ConvNeXt-T & Swin-N & SFKD 
& \textbf{77.83} 
& 29.80
& \textbf{11.58}
& \textbf{3.12}
& \textbf{20.51} 
& \textbf{1097.83} \\

\bottomrule
\end{tabular}%
}
\label{tabapp5}
\vspace{-0.3in}
\end{table*}

\subsection{Comparison Experiment of Training Computational Cost}
In addition to student performance improvement, training computational cost is also an important criterion for evaluating the practicality of knowledge distillation methods. 
Therefore, in~\cref{tabapp5}, we compare the training cost of SFKD with the conventional feature-based distillation method FitNet~\cite{romero2014fitnets} and the heterogeneous distillation method OFA~\cite{hao2023one}. 
The evaluated metrics include extra training parameters, total training FLOPs, extra training FLOPs, peak memory usage, and average Runtime per epoch. 
We conduct the comparison on three different teacher--student pairs on ImageNet-1K, as reported in~\cref{tabapp5}. 
The results show that SFKD achieves the best performance while maintaining competitive computational and memory costs, demonstrating favorable training efficiency and practical applicability.
\section{Conclusion and Future Work}
\label{appendix4}
In this work, we propose a Spatial--Frequency Joint-Aware Heterogeneous Knowledge Distillation framework.
In contrast to previous heterogeneous distillation methods that directly discard or suppress spatial information in heterogeneous representations, SFKD combines wavelet transforms with Fourier transforms to explicitly decouple spatial information in heterogeneous representations and selectively emphasizes the informative components.
Additionally, building on multi-level wavelet transforms, we further refine the wavelet sub-bands and design a novel Dual-Stream Dual-Stage Spectral Refinement module, which is applicable across different heterogeneous models.
By leveraging the complementary modeling capabilities of the different streams, DS$^2$SR effectively refines the local and global information in the wavelet sub-bands.
Finally, SFKD exploits the Fourier frequency domain’s ability to characterize global energy distribution and structural patterns, designing a Gaussian-filtered frequency loss function to selectively emphasize useful information in each sub-band.
To the best of our knowledge, this work is the first to integrate wavelet transforms and Fast Fourier transforms into heterogeneous knowledge distillation, demonstrating that spatial information in heterogeneous representations is not entirely redundant.
Experimental results show that effectively utilizing spatial information in representations can significantly enhance student model performance. 
The importance of further refining wavelet sub-bands and the efficiency of the DS$^2$SR module are also verified through experiments.

We note that the current refinement module may suffer from redundancy due to an excess of parameters, which can impact training efficiency.
Future work will focus on designing higher-performance refinement modules, while also combining them with lightweight modules to improve both performance and training efficiency.
We argue that our work provides a new direction for heterogeneous distillation through the integration of wavelet transforms and Fourier transforms.
Future work could further optimize this approach to achieve better knowledge transfer.




\end{document}